\documentclass[10pt,twocolumn,letterpaper]{article}

\usepackage{lipsum}
\usepackage{kotex}
\usepackage{multirow}
\usepackage{booktabs}
\usepackage[table]{xcolor}

\definecolor{mygreen}{HTML}{e2efda}
\newcommand{\greenrow}{\rowcolor{mygreen}}
\usepackage[pagenumbers]{iccv} 

\definecolor{iccvblue}{rgb}{0.21,0.49,0.74}
\usepackage[pagebackref,breaklinks,colorlinks,allcolors=iccvblue]{hyperref}

\newcommand{\paragrapht}[1]{\vspace{-10pt}\paragraph{#1}}

\def\paperID{11151} 
\def\confName{ICCV}
\def\confYear{2025}

\title{Vid-CamEdit: Video Camera Trajectory Editing \\ with Generative Rendering from Estimated Geometry}

\author {
    Junyoung Seo\textsuperscript{\rm 1}\footnotemark[1], \;
    Jisang Han\textsuperscript{\rm 1}\footnotemark[1], \;
    Jaewoo Jung\textsuperscript{\rm 1}\footnotemark[1], \;
    Siyoon Jin\textsuperscript{\rm 1}, \;
    JoungBin Lee\textsuperscript{\rm 1}, \\
    Takuya Narihira\textsuperscript{\rm 2}, \;
    Kazumi Fukuda\textsuperscript{\rm 2}, \;
    Takashi Shibuya\textsuperscript{\rm 2}, \;
    Donghoon Ahn\textsuperscript{\rm 1}, \\
    Shoukang Hu\textsuperscript{\rm 2}, \;
    Seungryong Kim\textsuperscript{\rm 1}\footnotemark[2], \;
    Yuki Mitsufuji\textsuperscript{\rm 2,3}\footnotemark[2] \\ \\
    \textsuperscript{\rm 1}KAIST AI \hspace{5pt}
    \textsuperscript{\rm 2}Sony AI
    \hspace{5pt}
    \textsuperscript{\rm 3}Sony Group Corporation \\
    {\tt\small\ \href{https://cvlab-kaist.github.io/Vid-CamEdit}{https://cvlab-kaist.github.io/Vid-CamEdit}}
 \\
}

\begin{document}

\twocolumn[{%
\renewcommand\twocolumn[1][]{#1}%
\maketitle
\vspace{-10pt}
\begin{center}
    \centering
    \captionsetup{type=figure}
    \includegraphics[width=1\textwidth]{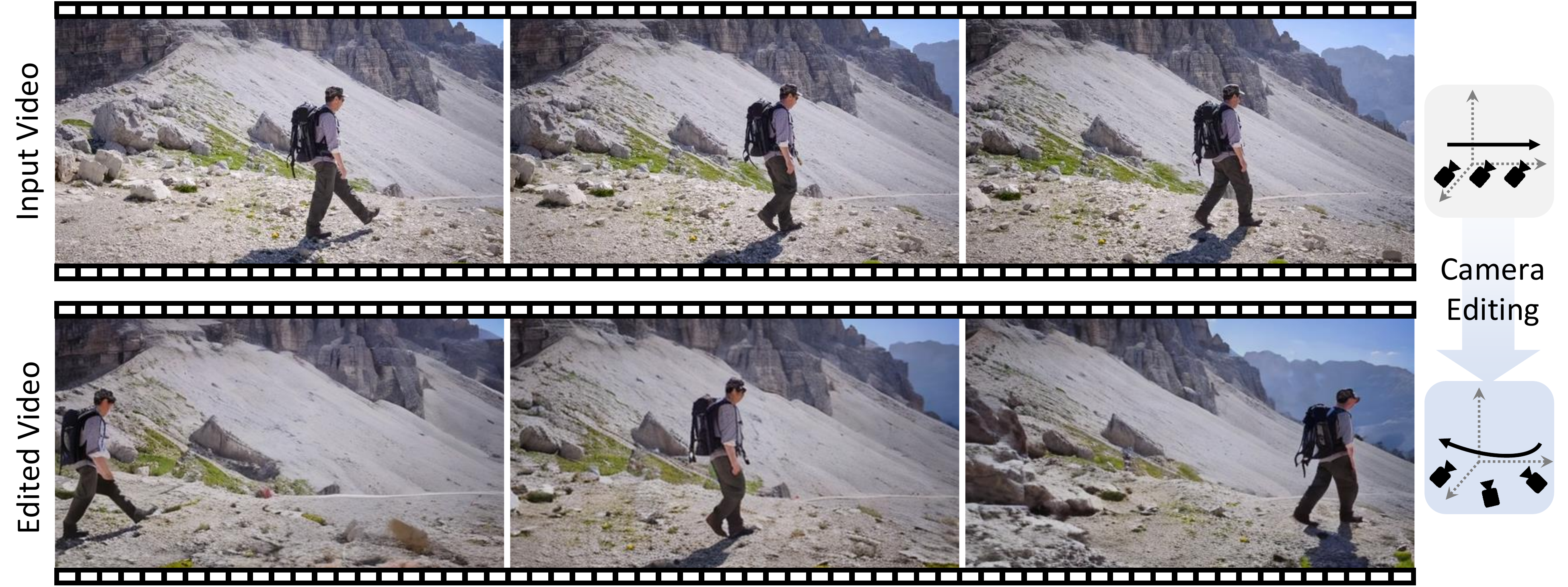}
    \vspace{-15pt}
    \captionof{figure}{\textbf{Teaser:} We aim to re-synthesize an input monocular video (top) following a desired camera trajectory. Our generated video (bottom) preserves the motion and structure of the input video while demonstrating realistic visual quality.}
    \label{fig:teaser}
\end{center}%
}]

\begin{abstract}
We introduce \textbf{Vid-CamEdit}, a novel framework for video camera trajectory editing, enabling the re-synthesis of monocular videos along user-defined camera paths. This task is challenging due to its ill-posed nature and the limited multi-view video data for training. Traditional reconstruction methods struggle with extreme trajectory changes, and existing generative models for dynamic novel view synthesis cannot handle in-the-wild videos. Our approach consists of two steps: estimating temporally consistent geometry, and generative rendering guided by this geometry. By integrating geometric priors, the generative model focuses on synthesizing realistic details where the estimated geometry is uncertain. We eliminate the need for extensive 4D training data through a factorized fine-tuning framework that separately trains spatial and temporal components using multi-view image and video data. Our method outperforms baselines in producing plausible videos from novel camera trajectories, especially in extreme extrapolation scenarios on real-world footage. 
\end{abstract}    
\vspace{-10pt}
\section{Introduction}
\label{sec:intro}
\begin{figure*}[t]
    \centering
    \includegraphics[width=\textwidth]{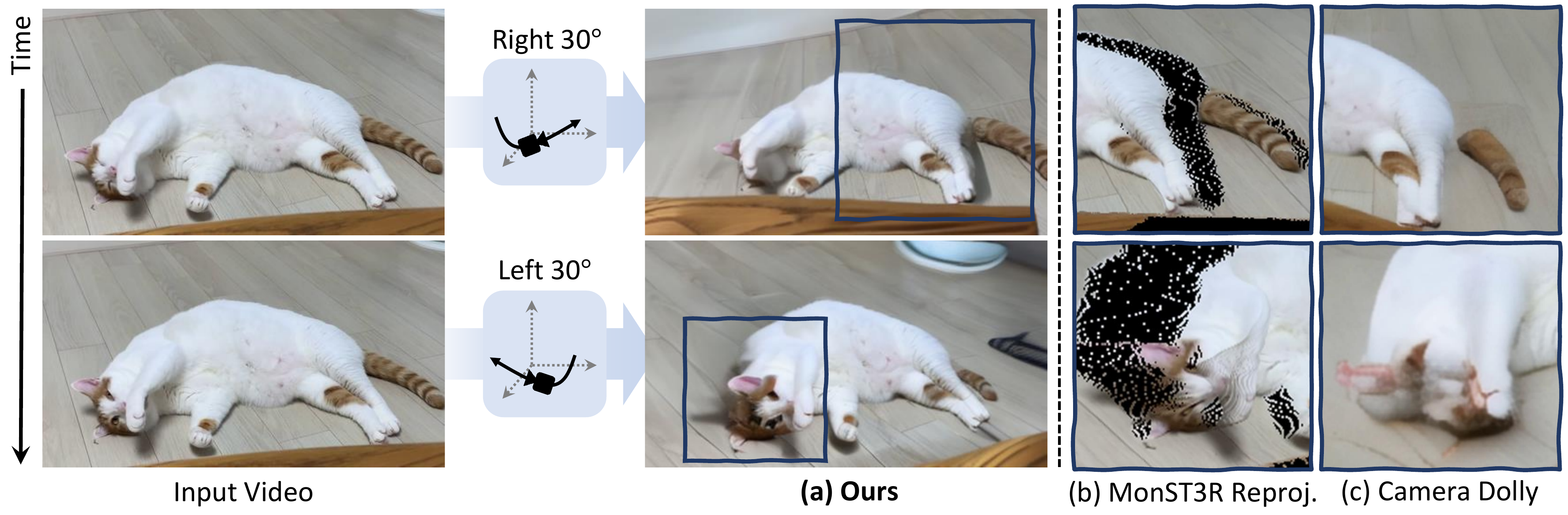} 
    \vspace{-15pt}
    \caption{\textbf{Motivation.} To edit camera trajectories in monocular videos, we embed knowledge from video geometry prediction models, \eg, MonST3R~\cite{zhang2024monst3r}, into video generative models~\cite{guo2023animatediff,blattmann2023stable}, allowing the model to synthesize realistic novel views by filling occluded regions the geometry model cannot infer. By incorporating geometrical cues for generation, our approach demonstrates superior performance on novel view video synthesis, compared to fully generative approaches \eg, Generative Camera Dolly~\cite{van2024generative}.}
    \vspace{-10pt}
    
    \label{fig:motivation}
\end{figure*}

When browsing through our camera albums, we often find ourselves wishing to view the videos we've captured from different camera poses. For instance, seeing footage originally shot from the side as if it were filmed from the front, or transforming a moving shot into one that appears as if taken from a stationary camera. \textit{What if we can freely manipulate the camera movement within recorded videos to re-synthesize them from any viewpoint?} This ability will not only revolutionize how we experience our own videos but also impact fields like video editing, 4D content creation~\cite{blattmann2023align,ho2022imagen,qi2023fatezero,teng2023drag}, virtual reality~\cite{fang2024vivid, stan2023ldm3d}, and robotics~\cite{chi2023diffusion,hoeg2024streamingdiffusionpolicyfast}.

In this work, we focus on the task of re-synthesizing a given video along a user-defined camera trajectory, allowing for arbitrary modifications to the camera's movement and perspective throughout the video, a process we refer to as \textit{video camera trajectory editing}. This task is inherently related to the extreme case of dynamic novel view synthesis (NVS) given a monocular video~\cite{li2023dynibar, wang2024shape}, as it involves generating views from significantly altered or entirely new camera trajectories that were not present in the original footage. Existing approaches encounter two main challenges when tackling this task:

\noindent \textbf{Reconstruction-based methods struggle with unseen areas.}
The extensive modification to the camera’s path makes the problem highly ill-posed, causing existing reconstruction-based methods for dynamic NVS~\cite{zhang2024monst3r, duan20244d, wu20244d, yang2024deformable, zhao2024pseudo} to fail in synthesizing visually realistic novel views, as illustrated in Fig.~\ref{fig:motivation}-(b). Because these methods focus on accurately reconstructing observed regions rather than handling unseen areas, they cannot accommodate the significant extrapolation required when the new camera trajectory deviates greatly from the original.

\noindent \textbf{Generation methods require large-scale 4D data.}
While generative models~\cite{ho2020denoising,rombach2022high,liu2023zero,sargent2023zeronvs,seo2024genwarp,gao2024cat3d} have shown promising results in synthesizing highly realistic novel views in static scenes by training on large-scale multi-view image datasets (3D data)~\cite{deitke2023objaverse,reizenstein2021common,zhou2018stereo}, applying this approach to dynamic scenes is challenging due to the limited availability of extensive real-world multi-view videos (4D data). Recent work, \eg, Generative Camera Dolly~\cite{van2024generative}, tackles the problem by training on synthetic multi-view video data. However, they often fail to generalize to real-world videos due to domain gaps, as shown in Fig.~\ref{fig:motivation}-(c).

In this work, to overcome these challenges in this task, we explore a more practical and data-efficient approach, called \textbf{Vid-CamEdit}, to leveraging generative models, which sidesteps the need for extensive real 4D training data. Instead of taking the data-driven solution, we decompose the task into two sub-tasks: (1) temporally-consistent geometry estimation and (2) generative video rendering based on the estimated geometry. Specifically, we ground pre-trained video generative models with geometry estimated from off-the-shelf geometry estimation models~\cite{zhang2024monst3r, hu2024depthcrafter, yang2024depth} (Fig.~\ref{fig:motivation}-(b)), allowing it to synthesize realistic novel view videos while relying on the geometry as a scaffold. This geometric prior reduces the burden on the generative model, enabling it to focus primarily on enhancing uncertain regions instead of learning full 4D dynamics from scratch, thereby greatly reducing the need for large-scale 4D training data.

To further reduce the need for 4D data, we incorporate a factorized fine-tuning strategy. By considering the spatio-temporal blocks of our video generative model independently, we train the spatial block with multi-view image (3D) data and train the temporal block with video data. As both 3D and video data are accessible up to scale, the training of generative models no longer requires 4D data.

Our main contributions are as follows:
\begin{itemize}
    \item We propose \textbf{Vid-CamEdit}, a geometric-grounded video-to-video translation framework to re-synthesize a user-provided video along a desired camera trajectory.
    \item By effectively grounding the estimated geometry to the video generative model and adopting a factorized fine-tuning strategy, our framework eliminates the need of large-scale 4D data for high-quality generation.
    \item Extensive experiments on Neu3D~\cite{li2022neural}, ST-NeRF~\cite{zhang2021editable}, and in-the-wild videos (\eg uncurated videos from online) validate that our framework achieves superior performance over existing methods~\cite{zhang2024monst3r,zhao2024pseudo, van2024generative}.
\end{itemize}
\section{Related Work}
\label{sec:relatedwork}
\paragraph{Dynamic novel view synthesis via reconstruction.}
Similar to traditional novel view synthesis, which aims to reconstruct the scene given multi-view observations, dynamic novel view synthesis extends its application to dynamic scenes. Building upon the success of Neural Radiance Fields (NeRF)~\cite{mildenhall2021nerf} and 3D Gaussian Splatting~\cite{kerbl20233d} for novel view synthesis, existing approaches~\cite{cao2023hexplane, fridovich2023k, ramasinghe2024blirf,duan20244d, wu20244d, yang2023real, yang2024deformable,wang2024shape,zhao2024pseudo, wang2024diffusion} tackle dynamic scenes by introducing an additional time-dimension~\cite{cao2023hexplane, fridovich2023k, ramasinghe2024blirf} or learning time-based deformations~\cite{duan20244d, wu20244d, yang2023real, yang2024deformable}. Although these approaches can effectively handle dynamics, due to the lack of ability to extrapolate or estimate unseen regions, novel views can only be obtained from viewpoints near the original input. These limitations hinder its application to in-the-wild videos, leading to large holes in unseen regions and accumulating reprojection errors.

\paragrapht{Video geometry estimation.}
Unlike monocular depth estimation (MDE)~\cite{yang2024depthany2, yang2024depthany, ke2024repurposing, ranftl2021vision, ranftl2020towards}, which infers depth from a single image, video depth estimation must ensure temporal consistency across frames. Early approaches~\cite{luo2020consistent, zhang2021consistent} achieved this by fine-tuning MDE models and modeling motions for each input video, refining depth predictions frame by frame. More recent methods~\cite{yang2024depth, hu2024depthcrafter, shao2024learning} leverage generative priors to enhance depth quality and stability. In parallel, a novel approach MonST3R~\cite{zhang2024monst3r} extends DUSt3R's~\cite{wang2024dust3r} unique pointmap representation, which capitalizes on accurate correspondence between images~\cite{hong2021deep,hong2022cost,hong2022neural,hong2024unifying2,cho2021cats,cho2022cats++}, enables dense 3D scene reconstruction to dynamic scenes by isolating dynamic and static regions. 

\paragrapht{Novel view generation in static scene.}
The advent of large-scale 3D object datasets such as Objaverse~\cite{deitke2023objaverse} has led to significant progress in novel view generative models for 3D objects~\cite{liu2023zero, liu2023syncdreamer, shi2023mvdream, wang2023imagedream, gao2024cat3d}. 
On the other hand, recent works~\cite{sargent2023zeronvs, seo2024genwarp, muller2024multidiff} have introduced scene-level novel view generative models based on multi-view scene datasets~\cite{zhou2018stereo, dai2017scannet, reizenstein2021common, liu2021infinite, liu2024reconx}. 
Compared to conventional reconstruction-based static NVS methods~\cite{mildenhall2021nerf,kerbl20233d,truong2023sparf,hong2024pf3plat,hong2024unifying}, they demonstrate superior extrapolation and interpolation abilities, particularly when input views are sparse. However, because these models are trained on multi-view image pairs as direct supervision, applying these methods directly to dynamic novel view synthesis requires multi-view video pairs, which are often difficult to obtain.

\paragrapht{Camera controllable video generation.}
Building upon the recent success of video diffusion models~\cite{blattmann2023stable, guo2023animatediff, yang2024cogvideox}, recent works~\cite{he2024cameractrl,xu2024camco,wang2024motionctrl, bahmani2024vd3d} have achieved camera-controllable video generation by introducing additional adapters into U-Net-based video diffusion models that accept camera trajectories. More recently, CVD~\cite{kuang2024collaborative} generates multi-view videos by equipping such camera-controllable video diffusion models with an attention-based synchronization module. However, all of these works only enable camera-conditioned video generation, whereas our goal is to tackle camera trajectory editing of the input video.  Concurrently, SynCamMaster~\cite{bai2024syncammaster} demonstrates more robust multi-view video generation using Unreal Engine–based synthetic data. However, it only generates stationary videos and has not been fully validated on user-provided video conditional generation.

\paragrapht{Generative dynamic novel view synthesis.}
Extending such camera-controllable video models to video-camera trajectory editing is non-trivial, as it requires both semantic understanding and low-level perception of the user-provided video.  Generative Camera Dolly~\cite{van2024generative} is the first attempt at this, paving the way for future research; however, it still shows clear weaknesses in generalizing to in-the-wild videos, being highly fitted to the 4D synthetic training data. 4DiM~\cite{watson2024controlling} generates novel view videos conditioned on one or more input images. However, it relies on 4D data from Google Street View and its generalizability to in-the-wild videos has not yet been validated. Recent concurrent efforts, such as ReCapture~\cite{zhang2024recapture} and CAT4D~\cite{wu2024cat4d}, share goals and motivations akin to ours. For instance, ReCapture enables this by leveraging LoRA for test-time training, whereas our approach generalizes without requiring any additional training at test time.
\section{Methodology}
\subsection{Problem definition}
 Given a monocular video as input, which can be captured from either a stationary or a moving camera, our objective of \textit{video camera trajectory editing} is to design a framework that can synthesize a new video from any desired camera trajectory.

 We first define the input video with $T$ frames of size $H \times W$ as $X \in \mathbb{R}^{T \times H \times W \times 3}$ and its camera trajectory as $C_X \in \mathbb{R}^{T \times 3 \times 4}$, which consists of a series of camera extrinsic matrices. The desired camera trajectory for the novel video $Y \in \mathbb{R}^{T \times H \times W \times 3}$ is defined as $C_Y \in \mathbb{R}^{T \times 3 \times 4}$, where $C_Y$ is obtained by applying per-frame relative camera transformations $C_\mathrm{rel} \in \mathbb{R}^{T \times 3 \times 4}$ to $C_X$. Altogether, our framework $\mathcal{F}(\cdot)$ synthesizes a new video $Y$ conditioned on the input video $X$ and relative camera transformations $C_\mathrm{rel}$ as follows:
\begin{equation}
    Y = \mathcal{F}(X, C_\mathrm{rel}, K),
\end{equation}
where we assume both the original and synthesized videos share the same camera intrinsics $K$.

To design this framework, the following conditions must be met: (1) The framework $\mathcal{F}(\cdot)$ should accept free-form camera trajectories, without being restricted to preset camera trajectories. (2) The synthesized video along the new camera trajectory $Y$ should preserve the geometric structure of the original video $X$. (3) The synthesized video $Y$ should appear visually realistic, with proper interpolation and extrapolation of regions that are not observed in the original video (\eg, occlusion areas).

\subsection{Overview and motivation}
To handle extensive extrapolation inherently required for our task, we design $\mathcal{F}(\cdot)$ as a generative framework, which has shown promising results in large extrapolation in static scenes~\cite{liu2023zero,sargent2023zeronvs}. However, leveraging generative models for dynamic NVS raises a unique challenge, which is the lack of sufficient real 4D data (multi-view videos). 

\begin{figure}[t]
    \centering
    \includegraphics[width=0.49\textwidth]{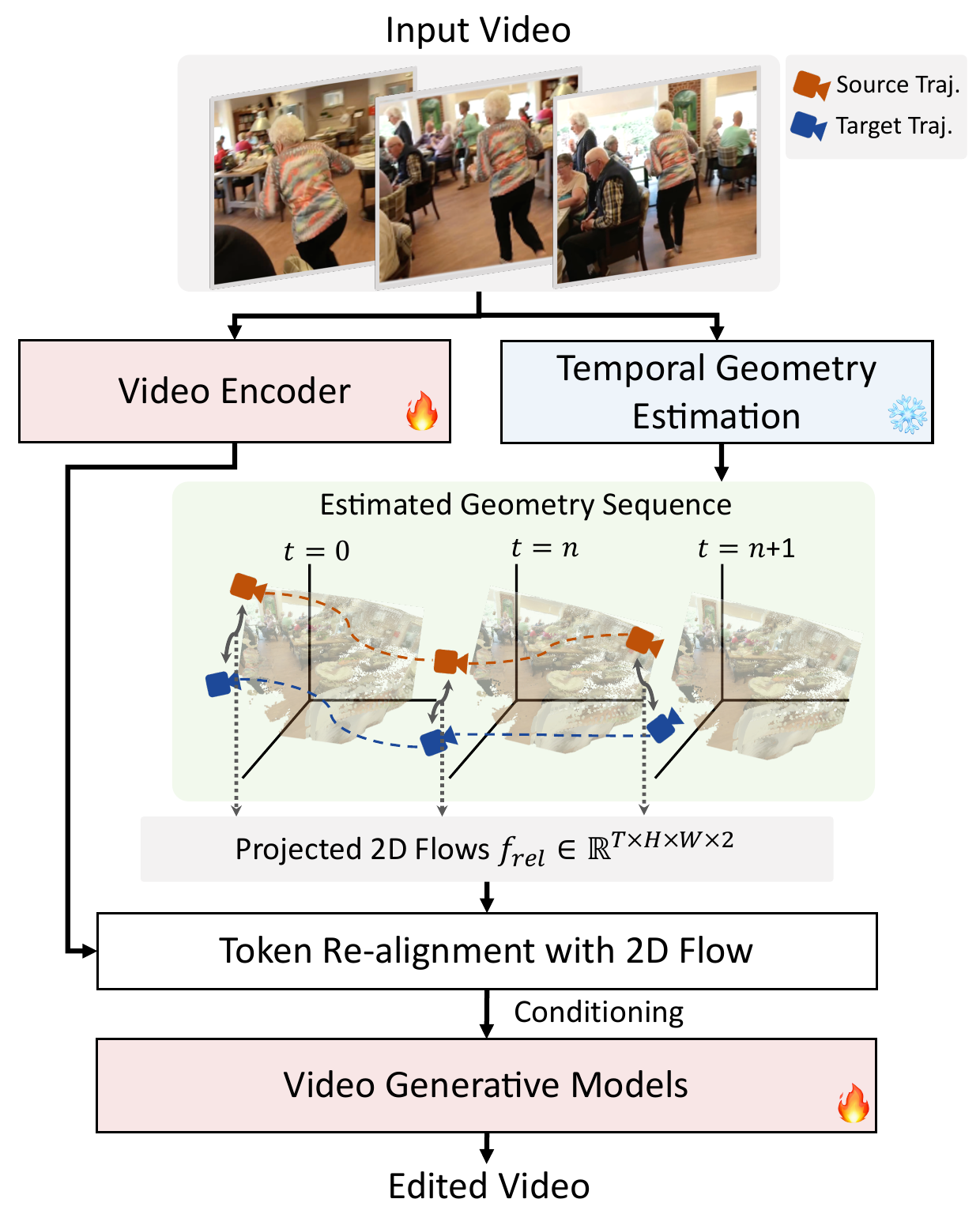} 
    \vspace{-10pt}
    \caption{\textbf{Overview of our framework.} Given a video and a target camera trajectory, we first extract video feature tokens using a Video Encoder and obtain the dynamic scene’s temporally consistent geometry through Temporal Geometry Estimation. We then ground the video generative model on this estimated geometry by re-aligning the video feature tokens according to the 2D flow between the source and target camera trajectories.}
    \vspace{-15pt}
    \label{fig:overview}
\end{figure}

To address this challenge, we explore a practical and data-efficient solution for the framework $\mathcal{F}(\cdot)$: a hybrid strategy that grounds strong geometry priors into video generative models. Our key intuition is to reduce the burden on the generative model by simplifying its task. Instead of relying solely on the generative model, we decompose the 4D problem into 3D spatial geometry and 1D temporal dynamics. For the 3D spatial geometry, we utilize a temporally consistent geometry estimation model to capture the 3D structure. As illustrated in Fig.~\ref{fig:motivation}-(b), this provides geometric cues to the video generation model, where the video generation model can utilize the geometry as a scaffold for realistic generation. To handle the 1D temporal dynamics, we leverage the temporal consistency capabilities inherent in video generative models. By exploiting these capabilities, we ensure that the generated frames are temporally coherent, preserving motion consistency over time.
The overview of our pipeline is illustrated in Fig.~\ref{fig:overview}.

\subsection{Generative rendering from estimated geometry}
\label{sec:method_rendering}

\paragraph{Temporally-consistent geometry estimation.}
To effectively reduce the burden on the video generative model of our framework $\mathcal{F}(\cdot)$, the geometric prediction model $g$ serves as a general model that is capable of estimating temporally consistent geometry of the given video. Although various models can be leveraged as $g$~\footnote{ Details on the formulation of other geometry estimation models~\cite{hu2024depthcrafter, yang2024depth, yang2024depthany2} (\eg, video depth models) can be found in Appendix~\ref{sec:appendix_geometry}.}, we build up our framework on the recently proposed MonST3R~\cite{zhang2024monst3r}, as its joint estimation of consistent camera trajectory and pointmaps can be effectively utilized in our framework. Specifically, the geometry of the input video is represented as a series of pointmaps $G \in \mathbb{R}^{T \times H \times W \times 3}$, which are coordinate maps indicating the 3D location of each pixel within the global 3D space. For each frame $t$, the pointmap $G_t$ provides a dense mapping from 2D pixel coordinates $(u,v)$ to their corresponding 3D world coordinates.

\paragrapht{Geometry-grounded video-to-video translation.}
With the estimated temporally-consistent geometry, we now reformulate the video generative model in our framework as a geometry-guided video-to-video translation problem. We incorporate the predicted geometry $G$ as a crucial cue alongside the desired camera trajectory. At a high level, this framework can be expressed as: 
\begin{equation}
\label{eq:sample}
\mathcal{F}(X, C_\mathrm{rel}, K) := \mathrm{Sample}\big( p_\theta(Y \mid X, C_\mathrm{rel}, K, G) \big),
\end{equation} 
where $p_\theta$ is a learned distribution of a diffusion model $\theta$ and $\mathrm{Sample}(\cdot)$ is a sampling function for diffusion reverse process~\cite{ho2020denoising, song2020denoising}. Although providing the 3D geometry information can facilitate novel view video generation, the model would still need to learn a mechanism that enables NVS that well reflects the input 3D geometry and camera parameters. Assuming that the pre-trained video generative model lacks the ability to explicitly understand 3D representations, we further simplify the task for the video model.

Given the pointmap $G_t$ for frame $t$, we can obtain 2D flow fields $f_\mathrm{rel}\in\mathbb{R}^{T \times H \times W \times 2}$ by projecting these 3D points onto the target viewpoint. Specifically, for each pixel $(u,v)$ in the source frame, we obtain the 2D flow $f_\mathrm{rel}$:
\begin{equation}
f_\mathrm{rel}(u,v,t) = \Pi(C_\mathrm{rel}(t) \cdot G(u,v,t), K) - (u,v),
\end{equation}
where $\Pi(\cdot)$ is the perspective projection function.

This process maps each source pixel to its corresponding location in the target view for the time $t$, effectively grounding the translation in geometry without requiring the model to handle and understand complex 3D structures directly. We thus reformulate the generative process as:
\begin{equation}
\mathcal{F}(X,C_\mathrm{rel},K) := \mathrm{Sample}\big( p_\theta(Y \mid X, f_\mathrm{rel}) \big{)},
\end{equation}
where we note that the previous conditions -- camera poses $C_\mathrm{rel}$, intrinsics $K$, and 3D geometry $G$ -- are all inherently embedded within the 2D flow maps $f_\mathrm{rel}$. This reformulation simplifies the task for the generative model while maintaining geometric consistency through the explicitly computed correspondences.

\paragrapht{Re-aligning input video tokens.}
\label{paragrpah:plucker}
For the reformulated video generative model that takes the input video and 2D flow maps as conditions, we incorporate both conditions into a pretrained video diffusion model~\cite{guo2023animatediff}. For video conditioning, the model must preserve the input video's details, such as color and texture. We adopt the architecture of ReferenceNet~\cite{hu2024animate}, which has been shown to effectively preserve the low-level semantics of input images~\cite{hu2024animate, men2024mimo}. Specifically, our approach is based on a U-Net-based video diffusion model where spatial and temporal blocks are interleaved. On top of this, we define a video encoder $\mathcal{E}_\phi$ that shares the same architecture as the video diffusion model. The feature tokens of $\mathcal{E}_\phi$ are then concatenated into the self-attention map of each spatial block in the diffusion model, which is for spatial interaction within each frame of the novel view video being generated.

For flow conditioning, we align the feature tokens of $\mathcal{E}_\phi$ with the flow condition $f_\mathrm{rel}$. To this end, we can either explicitly warp the feature tokens of input video~\cite{muller2024multidiff, niu2024mofa} or encourage the model to perform reliable internal re-alignment with flow-conditioning methods~\cite{seo2024genwarp, zhang2024tora, cai2024generative}. In this work, motivated by GenWarp~\cite{seo2024genwarp}, we rearrange the positional embeddings for input video according to the flow map $f_\mathrm{rel}$ and employ them as additional positional embeddings, thereby allowing the model to naturally learn the flow condition. Specifically, for a given position $(u,v)$ in frame $t$, the re-aligned positional encoding $\text{PE}'$ is computed as:
\begin{equation}
\text{PE}'(u,v,t) = \text{PE}\big(u + f_\mathrm{rel}(u,v,t)_x, v + f_\mathrm{rel}(u,v,t)_y, t \big),
\end{equation}
where $\text{PE}$ denotes the sinusoidal positional encoding for the input video, and $f_\mathrm{rel}(u,v)_x$, $ f_\mathrm{rel}(u,v)_y$ represents the flow vectors in $x$ and $y$ directions respectively.

The re-aligned positional embeddings are additionally incorporated into the video diffusion model alongside the original positional embeddings, enabling the video generative model to take the flow condition with the input video. Note that our re-alignment process is fully differentiable, allowing the overall framework to be trained end-to-end. Further details are provided in Appendix~\ref{sec:appendix_geometry}.

\begin{figure}[t]
    \centering
    \includegraphics[width=0.47\textwidth]{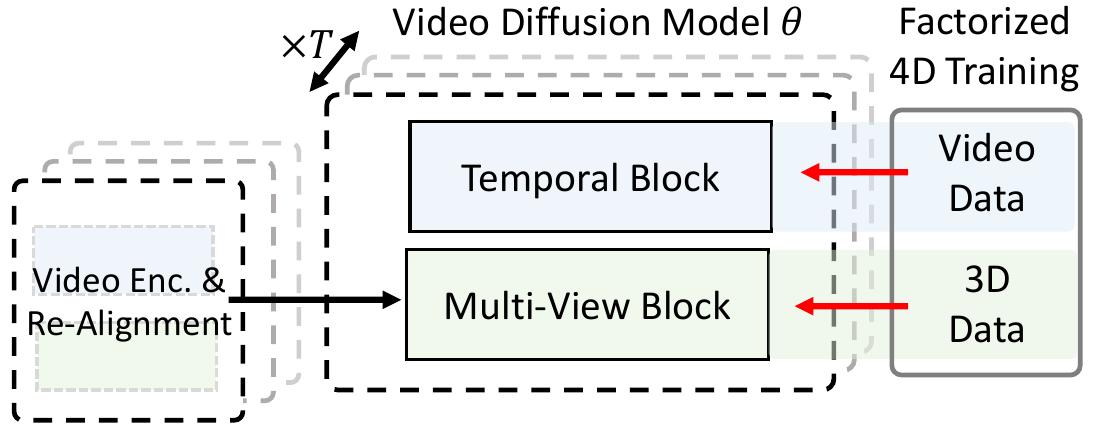} 
    \vspace{-8pt}
    \caption{\textbf{Factorized fine-tuning strategy without 4D data.}}
    \vspace{-12pt}
    \label{fig:training}
\end{figure}

\subsection{Fine-tuning without 4D data}
\label{sec:method_training}
While our geometry-grounded strategy effectively alleviates the computational burden on generative models, na\"ively training such models still hinges on real-world 4D data (\ie, multi-view videos), which is prohibitively expensive and impractical to acquire at large scales. Furthermore, training generative models exclusively on synthetic 4D data suffers from a substantial domain gap~\cite{van2024generative}, rendering it suboptimal. 
\begin{figure*}[t]
    \centering
    \includegraphics[width=0.9\textwidth]{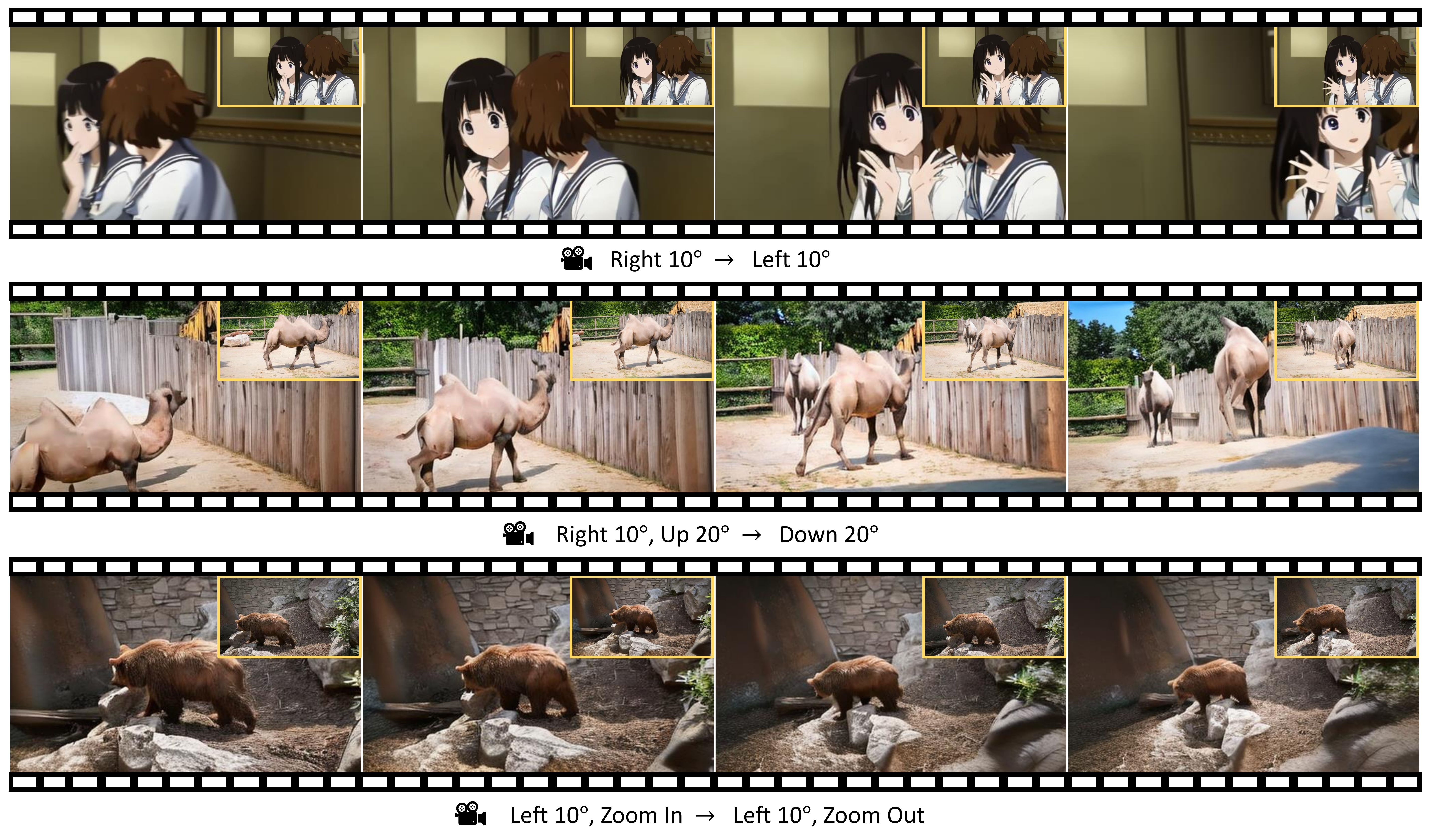} 
    \vspace{-10pt}
    \caption{\textbf{Qualitative results.} Given a user-provided monocular video, our method can synthesize high-quality videos along desired camera trajectories. The frames from the original videos are depicted in the yellow box of the top right of each image.}
    \vspace{-10pt}
    \label{fig:qualitative_results}
\end{figure*}

We instead adopt a factorized training protocol that capitalizes on more readily available datasets: multi-view images (3D) and conventional video data, similarly to~\cite{shao2024human4dit, watson2024controlling}. This shift obviates the need for comprehensive 4D data collection, offering a more scalable solution.

\paragrapht{Architecture.}
We employ a video generative model backbone~\cite{guo2023animatediff, blattmann2023stable} composed of interleaved spatial and temporal interaction blocks. We inject conditioning derived from input video tokens solely into the spatial interaction blocks -- thereby converting them into multi-view blocks (see Fig.~\ref{fig:training}) -- to concentrate on 3D synthesis. Meanwhile, the temporal interaction blocks remain dedicated to learning temporal priors. For detailed diagrams, please refer to Appendix~\ref{sec:appendix_geometry}.

\paragrapht{Block-wise supervision.}
Given our architectural design, we employ an intuitive factorized training strategy. When training on videos, we freeze the multi-view blocks; and when training on multi-view images, we freeze the temporal blocks. Multi-view images are treated as multi-view videos with $T=1$, updating only the multi-view blocks, whereas video data are treated as multi-view videos with the same input and output cameras, updating only the temporal blocks. Here, the conditioning tokens from the video encoder are replaced with a null condition at a predefined probability, similarly to CFG~\cite{ho2022classifier}.  By alternately freezing these blocks, we mitigate overfitting to either modality and successfully train our model without relying on 4D data.
\section{Experiments}
\subsection{Implementation details}
Our framework consists of two key components, geometry prediction model and video generative model. As mentioned in Section~\ref{sec:method_rendering}, we leverage MonST3R~\cite{zhang2024monst3r} as our geometry prediction model. For the video generative model, our framework can leverage any spatio-temporally factorized video diffusion models~\cite{guo2023animatediff, blattmann2023stable, chen2023videocrafter1}. Among them, we adopt AnimateDiff~\cite{guo2023animatediff} based on Stable Diffusion 1.5~\cite{rombach2022high} as our base model, generating $T=12$ frames at once, as it best fits our computational constraints. To condition the diffusion model with only the input video and cameras (2D flow), we replace the original text condition with CLIP~\cite{radford2021learning} image features. Please refer to Appendix~\ref{sec:appendix_impl} for additional details. Code and weights will be publicly available.

\begin{figure*}[t]
    \centering
    \includegraphics[width=0.95\textwidth]{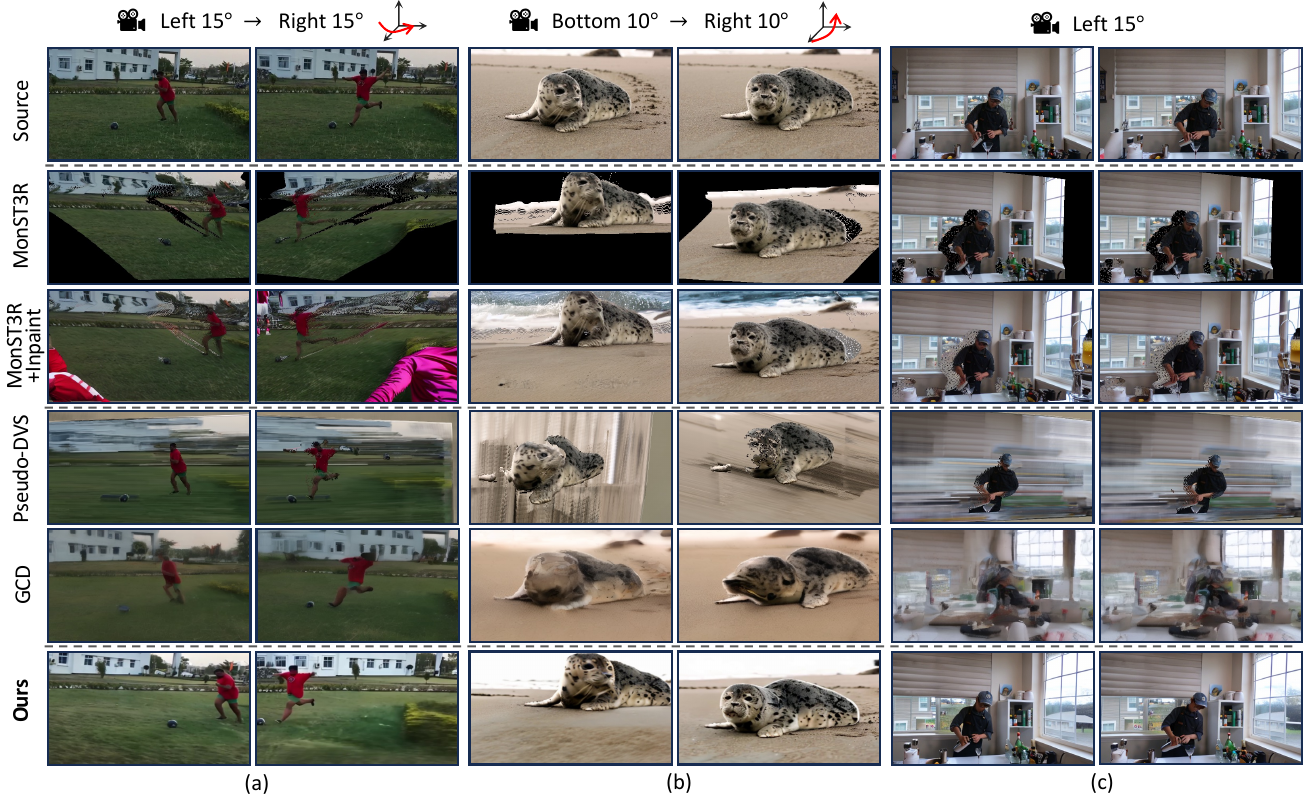}
    \vspace{-10pt}
    \caption{\textbf{Qualitative comparisons} with MonST3R~\cite{zhang2024monst3r}, video inpainting~\cite{zi2024cococo} w/MonST3R, GCD (Generative Camera Dolly)~\cite{van2024generative}, and Pseudo-DVS~\cite{zhao2024pseudo}. Ours is best in synthesizing visually realistic images while maintaining the original geometry.
    }
    \label{fig:qualitative_comp}
    \vspace{-5pt}
\end{figure*}
\begin{table*}[t]
    \small
    \centering
    \resizebox{0.95\linewidth}{!}{
    \begin{tabular}{l|cccc|cccc}
        \toprule
        \multirow{2}{*}{Methods} & \multicolumn{4}{c|}{\cellcolor{gray!20}Neu3D~\cite{li2022neural} } & \multicolumn{4}{c}{\cellcolor{gray!20}ST-NeRF dataset~\cite{zhang2021editable}}   \\


         & LPIPS $\downarrow$ & SSIM $\uparrow$ &  PSNR $\uparrow$ & Frame-Con. $\uparrow$ & LPIPS $\downarrow$ & SSIM $\uparrow$ & PSNR $\uparrow$ & Frame-Con. $\uparrow$   \\
        \midrule
        MonST3R~\cite{zhang2024monst3r} & 0.562 & 0.206 & 10.42 & 0.747 & 0.649 & 0.224 & 8.39 & 0.757 \\
        MonST3R~\cite{zhang2024monst3r} (Per-frame proj.) & \underline{0.453} & 0.291 & 11.73 & \underline{0.800} & 0.478 & 0.288 & 9.91 & \underline{0.811}  \\
        Pseudo-DVS~\cite{zhao2024pseudo} & 0.564 & \underline{0.352} & 14.43 & 0.655 & 0.527 & \textbf{0.415} & \textbf{15.33} & 0.742  \\
        Generative Camera Dolly~\cite{van2024generative} & 0.505 & 0.249 & 10.71 & 0.682 & \underline{0.425} & 0.346 & 13.60 & 0.748 \\

        
        \greenrow \textbf{Ours} & \textbf{0.414} & \textbf{0.358} & \textbf{14.91} & \textbf{0.858}&  \textbf{0.386} & \underline{0.381} & \underline{14.89} & \textbf{0.917} \\
        \bottomrule
    \end{tabular}
    }
    \caption{\textbf{Quantitative results with generalized/generation baselines.} We show quantitative comparisons in multi-view dynamic datasets, Neu3D~\cite{li2022neural} and ST-NeRF~\cite{zhang2021editable}. Ours is best in both LPIPS and Frame-Consistency (Frame-Con.), \ie, CLIP score between each frame of the input video and the generated video. Note that datasets containing videos from a stationary camera are used, as our primary baseline, Camera Dolly, does not officially accept input videos from moving cameras, e.g., DyCheck~\cite{gao2022monocular}.}
    \label{tab:main_quan}
    \vspace{-10pt}
\end{table*}

\paragrapht{Training dataset.}
For multi-view image data, we utilize RealEstate10K~\cite{zhou2018stereo}, Mannequin-Challenge~\cite{li2019learning}, MegaScene~\cite{tung2025megascenes}, and ScanNet~\cite{dai2017scannet}. For temporal fine-tuning, we initialize the temporal modules from the pre-trained checkpoint~\cite{guo2023animatediff} trained on WebVid-10M~\cite{bain2021frozen} and additionally use the TikTok dataset~\cite{jafarian2021learning} in fine-tuning.

\subsection{Baselines}
As our task demands extensive interpolation and extrapolation, we primarily compare our method with generation and generalizable methods: Generative Camera Dolly (GCD)~\cite{van2024generative} and Pseudo-DVS~\cite{zhao2024pseudo}. We also report performance improvements over our baseline: reprojection using MonST3R~\cite{zhang2024monst3r}.  We evaluate two variants with MonST3R: all-frame reprojection and per-frame reprojection. When synthesizing a novel view from frame $t$, all-frame reprojection leverages all pointmaps and dynamic masks from MonST3R's global alignment, projecting all static points across frames $[1, T]$ and combining them with dynamic points from frame $t$. 

 Furthermore, to validate our method’s applicability in per-scene 4D reconstruction scenarios, we integrate it into the existing optimization-based framework and compare its performance against Shape of Motions~\cite{wang2024shape},  DyniBar~\cite{li2023dynibar}, and HyperNeRF~\cite{park2021hypernerf}.

\subsection{Results}
\label{subsec:results}
\paragraph{Qualitative comparisons.}
Fig.~\ref{fig:qualitative_results} and Fig.~\ref{fig:qualitative_comp} show qualitative results and comparisons on in-the-wild videos with the baseline methods~\cite{zhang2024monst3r, zhao2024pseudo, van2024generative}. MonST3R~\cite{zhang2024monst3r} and Pseudo-DVS~\cite{zhao2024pseudo} show reasonable performance in some regions, however, failing to synthesize occluded regions. GCD~\cite{van2024generative} generates synthetic artifacts when dealing with in-the-wild video, failing to generalization. Additionally, we present results of reprojection-and-inpainting~\cite{zi2024cococo} baseline, struggling with refining ill-warped artifacts when conditioned on noisy reprojections.
In contrast, our method generates feasible videos from new camera trajectories. We also provide more qualitative results in Appendix~\ref{sec:appendix_exps}.

\paragrapht{Quantitative comparisons.}
We perform a quantitative comparison of our method and generalizable reconstruction and generation methods on the multi-view video datasets, Neu3D~\cite{li2022neural} and ST-NeRF dataset~\cite{zhang2021editable} in Tab.~\ref{tab:main_quan}. 

\begin{figure}[h]
    \centering
    \vspace{-7pt}
    \includegraphics[width=0.4\textwidth]{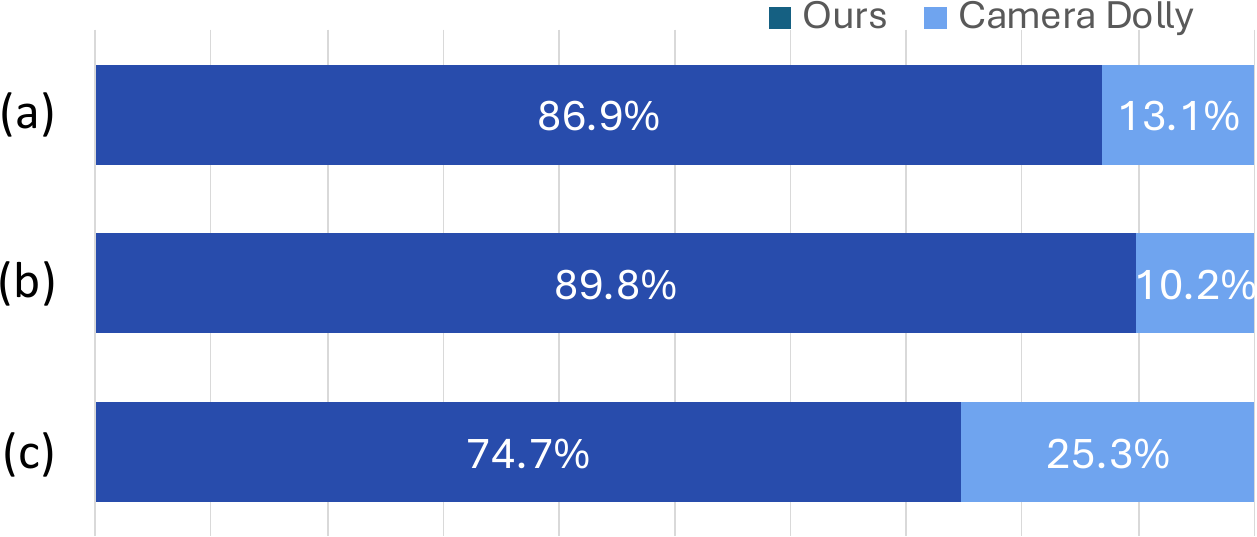} 

    \caption{\textbf{User study.} The user study is conducted by surveying 59 participants to evaluate (a) consistency to input videos, (b) video realness, and (c) faithfulness on camera trajectories.}
    \vspace{-10pt}
    
    \label{fig:userstudy}
\end{figure}

For frame consistency, we measure CLIP score between each frame of input videos and generated videos, following~\cite{jeong2024dreammotion}. The results show that our method achieves superior performance across all the datasets. 
Additionally, we report a user study for human preference in Fig.~\ref{fig:userstudy}, and  VBench~\cite{huang2024vbench} scores, VLM-based automated benchmarks in Fig.~\ref{fig:vbench}. Further details are described in Appendix~\ref{sec:appendix_userstudy}.

\paragrapht{Application for per-scene 4D reconstruction.}
While our primary goal is to directly generate video renderings, our approach can be seamlessly integrated into per-scene 4D reconstruction methods that produce 4D representations as output, by leveraging our generated results as additional supervision. As shown in Tab.~\ref{tab:recon_quan}, quantitative evaluations on the DyCheck dataset~\cite{gao2022monocular} demonstrate that incorporating our method into existing per-scene reconstruction pipelines yields higher reconstruction quality. Specifically, following Iterative Dataset Update proposed in~\cite{haque2023instruct}, we impose our iteratively generated results as additional training signals for Shape-of-Motion~\cite{wang2024shape}. Qualitative results and details can be found in Appendix~\ref{sec:appendix_additional_exps}.

\subsection{Analyses}
\begin{figure}[t]
    \centering
    \includegraphics[width=0.37\textwidth]{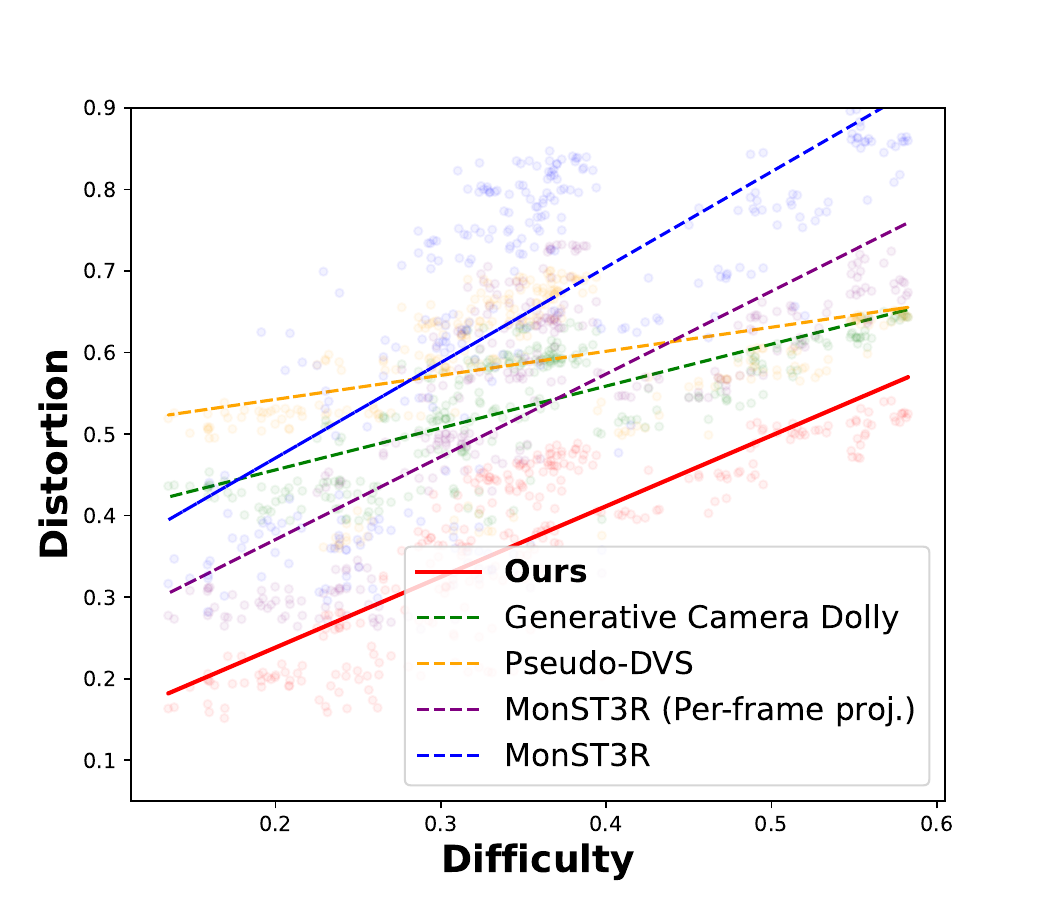} 
    \vspace{-5pt}
    \caption{\textbf{Comparison on various camera trajectory.} Following \cite{rombach2021geometry}, we measure LPIPS between generated videos and target videos (Distortion) over LPIPS between input videos and target videos (Difficulty). Ours consistently achieves best performance. }
    
    \label{fig:comp_difficulty}
\end{figure}

\paragraph{Performance on varying trajectory difficulties.}
We analyze how the performance of our method and the baseline methods changes as the desired camera trajectory becomes more challenging. Specifically, following the evaluation protocol in GeoGPT~\cite{rombach2021geometry}, we measure the LPIPS between the input video and the target GT video as the generation difficulty due to viewpoint change, and consider the LPIPS between the generated video and the target GT video as a degree of distortion. We then compare the degree of distortion against the generation difficulty. As shown in Fig.~\ref{fig:comp_difficulty}, our method demonstrates superior performance compared to the baseline methods as the difficulty increases.

\begin{figure}[t]
    \centering
    \includegraphics[width=0.42\textwidth]{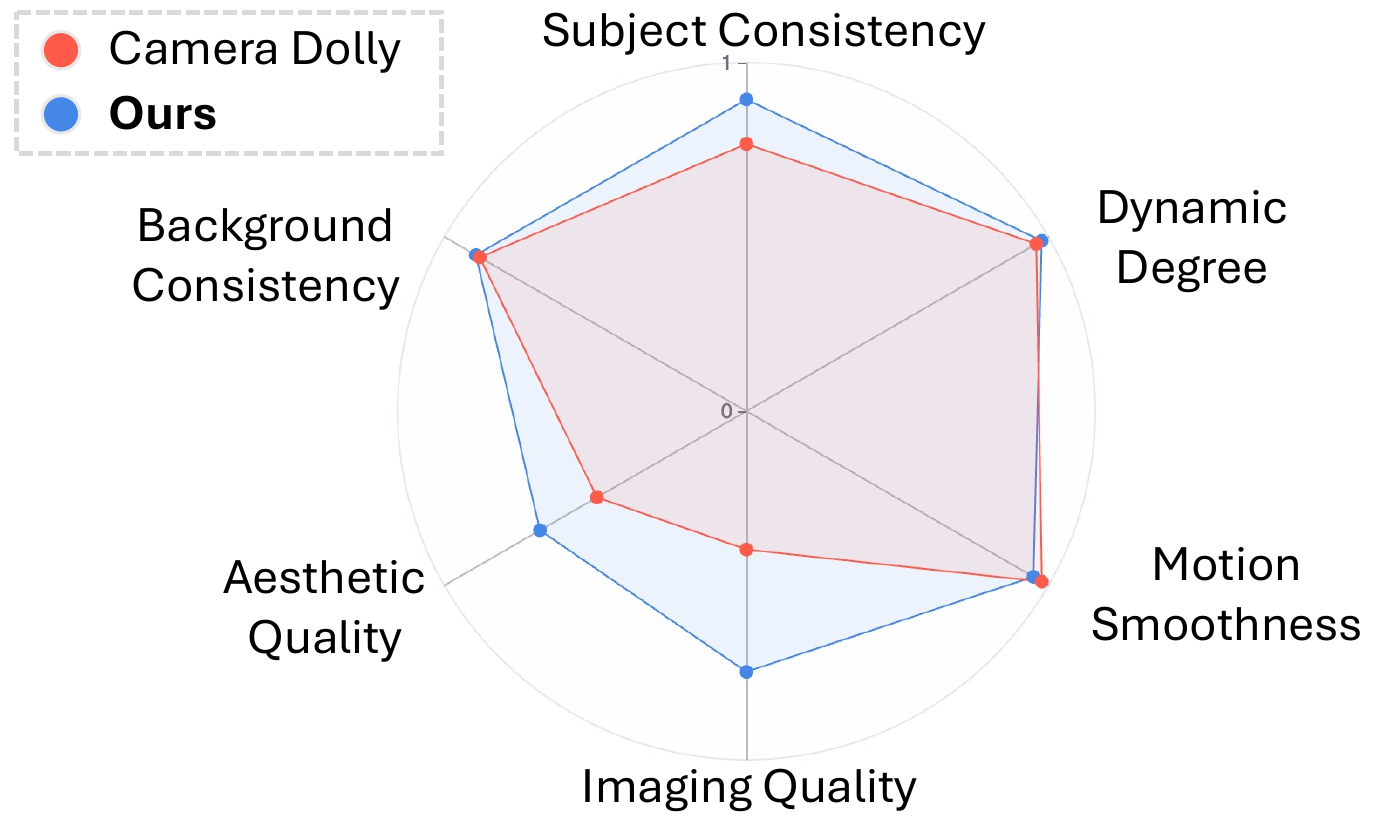} 
    \vspace{-5pt}
    \caption{\textbf{Quantitative comparisons on Vbench~\cite{huang2024vbench}} with Camera Dolly~\cite{van2024generative} on uncurated in-the-wild videos. 
    Our approach substantially outperforms Camera Dolly in both aesthetic and imaging quality.
    The large performance gap comes from Camera Dolly's synthetic-looking outputs.}
    
    \label{fig:vbench}
\end{figure}

\begin{table}[t]
    \centering
    \resizebox{1.0\linewidth}{!}{
        \begin{tabular}{l|ccc|ccc}
                \toprule
                \multirow{2}{*}{Method} 
                & \multicolumn{3}{c|}{\cellcolor{gray!20}Co-visible} 
                & \multicolumn{3}{c}{\cellcolor{gray!20}Occluded} \\
                & {PSNR $\uparrow$} & {LPIPS $\downarrow$} & {SSIM $\uparrow$} 
                & {PSNR $\uparrow$} & {LPIPS $\downarrow$} & {SSIM $\uparrow$} \\
                \midrule
                HyperNeRF~\cite{park2021hypernerf} & 15.99 & 0.510 & 0.590 
                          & - & - & - \\
                HyperNeRF* & 14.32 & 0.667 & 0.552
                          & 14.22 & 0.554 & 0.834 \\
                DyniBar~\cite{li2023dynibar} & 13.41 & 0.550 & 0.480
                        & - & - & - \\
                \midrule
                Shape-of-Motion~\cite{wang2024shape} & \underline{16.72} & 0.450 & \underline{0.630} 
                                 & -& -& -\\
                Shape-of-Motion* & 16.71 & 0.394 & \textbf{0.646}
                                 & \underline{15.24} & \underline{0.465} & \textbf{0.856}\\
                \greenrow \textbf{+ Ours} & \textbf{16.84} & \textbf{0.261} & 0.573
                              & \textbf{15.56} & \textbf{0.129} & \textbf{0.856}\\
                \bottomrule
        \end{tabular}}
    \vspace{-5pt}
    \caption{\textbf{Quantitative results of 4D per-scene reconstruction on DyCheck~\cite{gao2022monocular}.} Employing our method to the existing per-scene reconstruction method yield better reconstruction quality. $^*$~indicates the reproduced results for evaluation in occluded regions.}
    \vspace{-10pt}
    \label{tab:recon_quan}
\end{table}

\paragrapht{Ablation on design choices.}
We provide an ablation study on various design choices in our framework.
To validate the effectiveness of the geometry grounding introduced in Sec.~\ref{paragrpah:plucker}, we test a case where we directly inject camera poses (Plücker coordinates)~\cite{sitzmann2021light} into the model in place of this grounding. Additionally, we compare the performance against a baseline method: reprojection and video inpainting~\cite{zi2024cococo}, which is one possible na\"ive approach to combining geometry estimation models. As shown in Tab.~\ref{tab:ablation}, our full framework is most effective in both cases.

\paragrapht{Ablation on video geometry estimation models.}
We also provide an ablation study of using various video geometry estimation models in our framework. Specifically, we evaluate: MonST3R~\cite{zhang2024monst3r}, DepthAnyVideo~\cite{yang2024depth}, Depth-Anything2~\cite{yang2024depthany2}, and DepthCrafter~\cite{hu2024depthcrafter}. We evaluate the LPIPS and SSIM values on the Neu3D dataset~\cite{li2022neural}. As shown in Tab.~\ref{tab:abl_mde}, the results show minimal performance differences across models, indicating that our framework achieves consistent quality regardless of the chosen geometry prediction model.

\begin{table}[t]
\centering
\resizebox{0.4\textwidth}{!}{
    \begin{tabular}{l|cc}
        \toprule
        Baseline & LPIPS $\downarrow$ & SSIM $\uparrow$ \\
        \midrule
        w/o Geometry grounding  & 0.498 & 0.314 \\ 
        Reproj. + Inpainting~\cite{zi2024cococo} & 0.485 & 0.336 \\
        \greenrow \textbf{Ours} & \textbf{0.414} & \textbf{0.358} \\
        \bottomrule
    \end{tabular}
}
\caption{\textbf{Ablation study on design choices} comparing our framework with and without geometry grounding, substituting our framework with direct Plücker coordinate embedding~\cite{sitzmann2021light}, and a reprojection/video-inpainting baseline~\cite{zi2024cococo}.
Our full framework achieves the best performance.
\vspace{-10pt}
}
\label{tab:ablation}
\end{table}

\begin{table}[t]
\centering
\resizebox{0.38\textwidth}{!}{
    \begin{tabular}{l|cc}
    \toprule
    Base Geometry Models & LPIPS $\downarrow$ & SSIM $\uparrow$  \\
    \midrule
    DepthCrafter~\cite{hu2024depthcrafter} & \underline{0.416} & \underline{0.356} \\
    Depth-Anything 2~\cite{yang2024depthany2} & 0.420 & 0.354 \\
    DepthAnyVideo~\cite{yang2024depth} & 0.425 & 0.349 \\
    MonST3R~\cite{zhang2024monst3r} & \textbf{0.414} & \textbf{0.358} \\
    \bottomrule
    \end{tabular}
}
\caption{\textbf{Ablation on video geometry models.} We compare the performance of our framework leveraging different geometry prediction models $g$.}

\vspace{-10pt}
\label{tab:abl_mde}
\end{table}

\section{Conclusion}
We have introduced a novel framework \textbf{Vid-CamEdit} for video camera trajectory editing, using generative rendering grounded by estimated geometry. By combining temporally consistent geometry estimation and a factorized fine-tuning approach, we achieve robust and visually consistent novel view video synthesis.

\clearpage

{
    \small
    \bibliographystyle{ieeenat_fullname}
    \bibliography{main}

\begin{thebibliography}{112}
\providecommand{\natexlab}[1]{#1}
\providecommand{\url}[1]{\texttt{#1}}
\expandafter\ifx\csname urlstyle\endcsname\relax
  \providecommand{\doi}[1]{doi: #1}\else
  \providecommand{\doi}{doi: \begingroup \urlstyle{rm}\Url}\fi

\bibitem[Bahmani et~al.(2024)Bahmani, Skorokhodov, Siarohin, Menapace, Qian, Vasilkovsky, Lee, Wang, Zou, Tagliasacchi, et~al.]{bahmani2024vd3d}
Sherwin Bahmani, Ivan Skorokhodov, Aliaksandr Siarohin, Willi Menapace, Guocheng Qian, Michael Vasilkovsky, Hsin-Ying Lee, Chaoyang Wang, Jiaxu Zou, Andrea Tagliasacchi, et~al.
\newblock Vd3d: Taming large video diffusion transformers for 3d camera control.
\newblock \emph{arXiv preprint arXiv:2407.12781}, 2024.

\bibitem[Bai et~al.(2024)Bai, Xia, Wang, Yuan, Fu, Liu, Hu, Wan, and Zhang]{bai2024syncammaster}
Jianhong Bai, Menghan Xia, Xintao Wang, Ziyang Yuan, Xiao Fu, Zuozhu Liu, Haoji Hu, Pengfei Wan, and Di Zhang.
\newblock Syncammaster: Synchronizing multi-camera video generation from diverse viewpoints.
\newblock \emph{arXiv preprint arXiv:2412.07760}, 2024.

\bibitem[Bain et~al.(2021)Bain, Nagrani, Varol, and Zisserman]{bain2021frozen}
Max Bain, Arsha Nagrani, G{\"u}l Varol, and Andrew Zisserman.
\newblock Frozen in time: A joint video and image encoder for end-to-end retrieval.
\newblock In \emph{Proceedings of the IEEE/CVF international conference on computer vision}, pages 1728--1738, 2021.

\bibitem[Blattmann et~al.(2023{\natexlab{a}})Blattmann, Dockhorn, Kulal, Mendelevitch, Kilian, Lorenz, Levi, English, Voleti, Letts, et~al.]{blattmann2023stable}
Andreas Blattmann, Tim Dockhorn, Sumith Kulal, Daniel Mendelevitch, Maciej Kilian, Dominik Lorenz, Yam Levi, Zion English, Vikram Voleti, Adam Letts, et~al.
\newblock Stable video diffusion: Scaling latent video diffusion models to large datasets.
\newblock \emph{arXiv preprint arXiv:2311.15127}, 2023{\natexlab{a}}.

\bibitem[Blattmann et~al.(2023{\natexlab{b}})Blattmann, Rombach, Ling, Dockhorn, Kim, Fidler, and Kreis]{blattmann2023align}
Andreas Blattmann, Robin Rombach, Huan Ling, Tim Dockhorn, Seung~Wook Kim, Sanja Fidler, and Karsten Kreis.
\newblock Align your latents: High-resolution video synthesis with latent diffusion models.
\newblock In \emph{Proceedings of the IEEE/CVF Conference on Computer Vision and Pattern Recognition}, pages 22563--22575, 2023{\natexlab{b}}.

\bibitem[Cai et~al.(2024)Cai, Ceylan, Gadelha, Huang, Wang, and Wetzstein]{cai2024generative}
Shengqu Cai, Duygu Ceylan, Matheus Gadelha, Chun-Hao~Paul Huang, Tuanfeng~Yang Wang, and Gordon Wetzstein.
\newblock Generative rendering: Controllable 4d-guided video generation with 2d diffusion models.
\newblock In \emph{Proceedings of the IEEE/CVF Conference on Computer Vision and Pattern Recognition}, pages 7611--7620, 2024.

\bibitem[Cao and Johnson(2023)]{cao2023hexplane}
Ang Cao and Justin Johnson.
\newblock Hexplane: A fast representation for dynamic scenes.
\newblock In \emph{Proceedings of the IEEE/CVF Conference on Computer Vision and Pattern Recognition}, pages 130--141, 2023.

\bibitem[Caron et~al.(2021)Caron, Touvron, Misra, J{\'e}gou, Mairal, Bojanowski, and Joulin]{caron2021emerging}
Mathilde Caron, Hugo Touvron, Ishan Misra, Herv{\'e} J{\'e}gou, Julien Mairal, Piotr Bojanowski, and Armand Joulin.
\newblock Emerging properties in self-supervised vision transformers.
\newblock In \emph{Proceedings of the IEEE/CVF international conference on computer vision}, pages 9650--9660, 2021.

\bibitem[Chen et~al.(2023)Chen, Xia, He, Zhang, Cun, Yang, Xing, Liu, Chen, Wang, et~al.]{chen2023videocrafter1}
Haoxin Chen, Menghan Xia, Yingqing He, Yong Zhang, Xiaodong Cun, Shaoshu Yang, Jinbo Xing, Yaofang Liu, Qifeng Chen, Xintao Wang, et~al.
\newblock Videocrafter1: Open diffusion models for high-quality video generation.
\newblock \emph{arXiv preprint arXiv:2310.19512}, 2023.

\bibitem[Chi et~al.(2023)Chi, Xu, Feng, Cousineau, Du, Burchfiel, Tedrake, and Song]{chi2023diffusion}
Cheng Chi, Zhenjia Xu, Siyuan Feng, Eric Cousineau, Yilun Du, Benjamin Burchfiel, Russ Tedrake, and Shuran Song.
\newblock Diffusion policy: Visuomotor policy learning via action diffusion.
\newblock \emph{The International Journal of Robotics Research}, page 02783649241273668, 2023.

\bibitem[Cho et~al.(2021)Cho, Hong, Jeon, Lee, Sohn, and Kim]{cho2021cats}
Seokju Cho, Sunghwan Hong, Sangryul Jeon, Yunsung Lee, Kwanghoon Sohn, and Seungryong Kim.
\newblock Cats: Cost aggregation transformers for visual correspondence.
\newblock \emph{Advances in Neural Information Processing Systems}, 34:\penalty0 9011--9023, 2021.

\bibitem[Cho et~al.(2022)Cho, Hong, and Kim]{cho2022cats++}
Seokju Cho, Sunghwan Hong, and Seungryong Kim.
\newblock Cats++: Boosting cost aggregation with convolutions and transformers.
\newblock \emph{IEEE Transactions on Pattern Analysis and Machine Intelligence}, 45\penalty0 (6):\penalty0 7174--7194, 2022.

\bibitem[Dai et~al.(2017)Dai, Chang, Savva, Halber, Funkhouser, and Nie{\ss}ner]{dai2017scannet}
Angela Dai, Angel~X Chang, Manolis Savva, Maciej Halber, Thomas Funkhouser, and Matthias Nie{\ss}ner.
\newblock Scannet: Richly-annotated 3d reconstructions of indoor scenes.
\newblock In \emph{Proceedings of the IEEE conference on computer vision and pattern recognition}, pages 5828--5839, 2017.

\bibitem[Deitke et~al.(2023)Deitke, Schwenk, Salvador, Weihs, Michel, VanderBilt, Schmidt, Ehsani, Kembhavi, and Farhadi]{deitke2023objaverse}
Matt Deitke, Dustin Schwenk, Jordi Salvador, Luca Weihs, Oscar Michel, Eli VanderBilt, Ludwig Schmidt, Kiana Ehsani, Aniruddha Kembhavi, and Ali Farhadi.
\newblock Objaverse: A universe of annotated 3d objects.
\newblock In \emph{Proceedings of the IEEE/CVF Conference on Computer Vision and Pattern Recognition}, pages 13142--13153, 2023.

\bibitem[Duan et~al.(2024)Duan, Wei, Dai, He, Chen, and Chen]{duan20244d}
Yuanxing Duan, Fangyin Wei, Qiyu Dai, Yuhang He, Wenzheng Chen, and Baoquan Chen.
\newblock 4d gaussian splatting: Towards efficient novel view synthesis for dynamic scenes.
\newblock \emph{arXiv preprint arXiv:2402.03307}, 2024.

\bibitem[Edstedt et~al.(2024)Edstedt, Sun, B{\"o}kman, Wadenb{\"a}ck, and Felsberg]{edstedt2024roma}
Johan Edstedt, Qiyu Sun, Georg B{\"o}kman, M{\aa}rten Wadenb{\"a}ck, and Michael Felsberg.
\newblock Roma: Robust dense feature matching.
\newblock In \emph{Proceedings of the IEEE/CVF Conference on Computer Vision and Pattern Recognition}, pages 19790--19800, 2024.

\bibitem[Fang et~al.(2024)Fang, Zhai, Su, Song, Zhu, Wang, Chen, Liu, Cao, and Zha]{fang2024vivid}
Zixun Fang, Wei Zhai, Aimin Su, Hongliang Song, Kai Zhu, Mao Wang, Yu Chen, Zhiheng Liu, Yang Cao, and Zheng-Jun Zha.
\newblock Vivid: Video virtual try-on using diffusion models.
\newblock \emph{arXiv preprint arXiv:2405.11794}, 2024.

\bibitem[Fischler and Bolles(1981)]{fischler1981random}
Martin~A Fischler and Robert~C Bolles.
\newblock Random sample consensus: a paradigm for model fitting with applications to image analysis and automated cartography.
\newblock \emph{Communications of the ACM}, 24\penalty0 (6):\penalty0 381--395, 1981.

\bibitem[Fridovich-Keil et~al.(2023)Fridovich-Keil, Meanti, Warburg, Recht, and Kanazawa]{fridovich2023k}
Sara Fridovich-Keil, Giacomo Meanti, Frederik~Rahb{\ae}k Warburg, Benjamin Recht, and Angjoo Kanazawa.
\newblock K-planes: Explicit radiance fields in space, time, and appearance.
\newblock In \emph{Proceedings of the IEEE/CVF Conference on Computer Vision and Pattern Recognition}, pages 12479--12488, 2023.

\bibitem[Gao et~al.(2022)Gao, Li, Tulsiani, Russell, and Kanazawa]{gao2022monocular}
Hang Gao, Ruilong Li, Shubham Tulsiani, Bryan Russell, and Angjoo Kanazawa.
\newblock Monocular dynamic view synthesis: A reality check.
\newblock \emph{Advances in Neural Information Processing Systems}, 35:\penalty0 33768--33780, 2022.

\bibitem[Gao et~al.(2024)Gao, Holynski, Henzler, Brussee, Martin-Brualla, Srinivasan, Barron, and Poole]{gao2024cat3d}
Ruiqi Gao, Aleksander Holynski, Philipp Henzler, Arthur Brussee, Ricardo Martin-Brualla, Pratul Srinivasan, Jonathan~T Barron, and Ben Poole.
\newblock Cat3d: Create anything in 3d with multi-view diffusion models.
\newblock \emph{arXiv preprint arXiv:2405.10314}, 2024.

\bibitem[Greff et~al.(2022)Greff, Belletti, Beyer, Doersch, Du, Duckworth, Fleet, Gnanapragasam, Golemo, Herrmann, et~al.]{greff2022kubric}
Klaus Greff, Francois Belletti, Lucas Beyer, Carl Doersch, Yilun Du, Daniel Duckworth, David~J Fleet, Dan Gnanapragasam, Florian Golemo, Charles Herrmann, et~al.
\newblock Kubric: A scalable dataset generator.
\newblock In \emph{Proceedings of the IEEE/CVF conference on computer vision and pattern recognition}, pages 3749--3761, 2022.

\bibitem[Guo et~al.(2023)Guo, Yang, Rao, Liang, Wang, Qiao, Agrawala, Lin, and Dai]{guo2023animatediff}
Yuwei Guo, Ceyuan Yang, Anyi Rao, Zhengyang Liang, Yaohui Wang, Yu Qiao, Maneesh Agrawala, Dahua Lin, and Bo Dai.
\newblock Animatediff: Animate your personalized text-to-image diffusion models without specific tuning.
\newblock \emph{arXiv preprint arXiv:2307.04725}, 2023.

\bibitem[Haque et~al.(2023)Haque, Tancik, Efros, Holynski, and Kanazawa]{haque2023instruct}
Ayaan Haque, Matthew Tancik, Alexei~A Efros, Aleksander Holynski, and Angjoo Kanazawa.
\newblock Instruct-nerf2nerf: Editing 3d scenes with instructions.
\newblock In \emph{Proceedings of the IEEE/CVF International Conference on Computer Vision}, pages 19740--19750, 2023.

\bibitem[Hartley and Zisserman(2003)]{hartley2003multiple}
Richard Hartley and Andrew Zisserman.
\newblock \emph{Multiple view geometry in computer vision}.
\newblock Cambridge university press, 2003.

\bibitem[He et~al.(2024)He, Xu, Guo, Wetzstein, Dai, Li, and Yang]{he2024cameractrl}
Hao He, Yinghao Xu, Yuwei Guo, Gordon Wetzstein, Bo Dai, Hongsheng Li, and Ceyuan Yang.
\newblock Cameractrl: Enabling camera control for text-to-video generation.
\newblock \emph{arXiv preprint arXiv:2404.02101}, 2024.

\bibitem[Ho and Salimans(2022)]{ho2022classifier}
Jonathan Ho and Tim Salimans.
\newblock Classifier-free diffusion guidance.
\newblock \emph{arXiv preprint arXiv:2207.12598}, 2022.

\bibitem[Ho et~al.(2020)Ho, Jain, and Abbeel]{ho2020denoising}
Jonathan Ho, Ajay Jain, and Pieter Abbeel.
\newblock Denoising diffusion probabilistic models.
\newblock \emph{Advances in neural information processing systems}, 33:\penalty0 6840--6851, 2020.

\bibitem[Ho et~al.(2022)Ho, Chan, Saharia, Whang, Gao, Gritsenko, Kingma, Poole, Norouzi, Fleet, et~al.]{ho2022imagen}
Jonathan Ho, William Chan, Chitwan Saharia, Jay Whang, Ruiqi Gao, Alexey Gritsenko, Diederik~P Kingma, Ben Poole, Mohammad Norouzi, David~J Fleet, et~al.
\newblock Imagen video: High definition video generation with diffusion models.
\newblock \emph{arXiv preprint arXiv:2210.02303}, 2022.

\bibitem[Hong and Kim(2021)]{hong2021deep}
Sunghwan Hong and Seungryong Kim.
\newblock Deep matching prior: Test-time optimization for dense correspondence.
\newblock In \emph{Proceedings of the IEEE/CVF international conference on computer vision}, pages 9907--9917, 2021.

\bibitem[Hong et~al.(2022{\natexlab{a}})Hong, Cho, Nam, Lin, and Kim]{hong2022cost}
Sunghwan Hong, Seokju Cho, Jisu Nam, Stephen Lin, and Seungryong Kim.
\newblock Cost aggregation with 4d convolutional swin transformer for few-shot segmentation.
\newblock In \emph{European Conference on Computer Vision}, pages 108--126. Springer, 2022{\natexlab{a}}.

\bibitem[Hong et~al.(2022{\natexlab{b}})Hong, Nam, Cho, Hong, Jeon, Min, and Kim]{hong2022neural}
Sunghwan Hong, Jisu Nam, Seokju Cho, Susung Hong, Sangryul Jeon, Dongbo Min, and Seungryong Kim.
\newblock Neural matching fields: Implicit representation of matching fields for visual correspondence.
\newblock \emph{Advances in Neural Information Processing Systems}, 35:\penalty0 13512--13526, 2022{\natexlab{b}}.

\bibitem[Hong et~al.(2024{\natexlab{a}})Hong, Cho, Kim, and Lin]{hong2024unifying2}
Sunghwan Hong, Seokju Cho, Seungryong Kim, and Stephen Lin.
\newblock Unifying feature and cost aggregation with transformers for semantic and visual correspondence.
\newblock \emph{arXiv preprint arXiv:2403.11120}, 2024{\natexlab{a}}.

\bibitem[Hong et~al.(2024{\natexlab{b}})Hong, Jung, Shin, Han, Yang, Luo, and Kim]{hong2024pf3plat}
Sunghwan Hong, Jaewoo Jung, Heeseong Shin, Jisang Han, Jiaolong Yang, Chong Luo, and Seungryong Kim.
\newblock Pf3plat: Pose-free feed-forward 3d gaussian splatting.
\newblock \emph{arXiv preprint arXiv:2410.22128}, 2024{\natexlab{b}}.

\bibitem[Hong et~al.(2024{\natexlab{c}})Hong, Jung, Shin, Yang, Kim, and Luo]{hong2024unifying}
Sunghwan Hong, Jaewoo Jung, Heeseong Shin, Jiaolong Yang, Seungryong Kim, and Chong Luo.
\newblock Unifying correspondence pose and nerf for generalized pose-free novel view synthesis.
\newblock In \emph{Proceedings of the IEEE/CVF Conference on Computer Vision and Pattern Recognition}, pages 20196--20206, 2024{\natexlab{c}}.

\bibitem[Hu(2024)]{hu2024animate}
Li Hu.
\newblock Animate anyone: Consistent and controllable image-to-video synthesis for character animation.
\newblock In \emph{Proceedings of the IEEE/CVF Conference on Computer Vision and Pattern Recognition}, pages 8153--8163, 2024.

\bibitem[Hu et~al.(2024)Hu, Gao, Li, Zhao, Cun, Zhang, Quan, and Shan]{hu2024depthcrafter}
Wenbo Hu, Xiangjun Gao, Xiaoyu Li, Sijie Zhao, Xiaodong Cun, Yong Zhang, Long Quan, and Ying Shan.
\newblock Depthcrafter: Generating consistent long depth sequences for open-world videos.
\newblock \emph{arXiv preprint arXiv:2409.02095}, 2024.

\bibitem[Huang et~al.(2024)Huang, He, Yu, Zhang, Si, Jiang, Zhang, Wu, Jin, Chanpaisit, et~al.]{huang2024vbench}
Ziqi Huang, Yinan He, Jiashuo Yu, Fan Zhang, Chenyang Si, Yuming Jiang, Yuanhan Zhang, Tianxing Wu, Qingyang Jin, Nattapol Chanpaisit, et~al.
\newblock Vbench: Comprehensive benchmark suite for video generative models.
\newblock In \emph{Proceedings of the IEEE/CVF Conference on Computer Vision and Pattern Recognition}, pages 21807--21818, 2024.

\bibitem[Høeg et~al.(2024)Høeg, Du, and Egeland]{hoeg2024streamingdiffusionpolicyfast}
Sigmund~H. Høeg, Yilun Du, and Olav Egeland.
\newblock Streaming diffusion policy: Fast policy synthesis with variable noise diffusion models, 2024.

\bibitem[Jafarian and Park(2021)]{jafarian2021learning}
Yasamin Jafarian and Hyun~Soo Park.
\newblock Learning high fidelity depths of dressed humans by watching social media dance videos.
\newblock In \emph{Proceedings of the IEEE/CVF Conference on Computer Vision and Pattern Recognition}, pages 12753--12762, 2021.

\bibitem[Jeong et~al.(2024)Jeong, Chang, Park, and Ye]{jeong2024dreammotion}
Hyeonho Jeong, Jinho Chang, Geon~Yeong Park, and Jong~Chul Ye.
\newblock Dreammotion: Space-time self-similar score distillation for zero-shot video editing.
\newblock \emph{arXiv preprint arXiv:2403.12002}, 2024.

\bibitem[Ke et~al.(2024)Ke, Obukhov, Huang, Metzger, Daudt, and Schindler]{ke2024repurposing}
Bingxin Ke, Anton Obukhov, Shengyu Huang, Nando Metzger, Rodrigo~Caye Daudt, and Konrad Schindler.
\newblock Repurposing diffusion-based image generators for monocular depth estimation.
\newblock In \emph{Proceedings of the IEEE/CVF Conference on Computer Vision and Pattern Recognition}, pages 9492--9502, 2024.

\bibitem[Ke et~al.(2021)Ke, Wang, Wang, Milanfar, and Yang]{ke2021musiq}
Junjie Ke, Qifei Wang, Yilin Wang, Peyman Milanfar, and Feng Yang.
\newblock Musiq: Multi-scale image quality transformer.
\newblock In \emph{Proceedings of the IEEE/CVF international conference on computer vision}, pages 5148--5157, 2021.

\bibitem[Kerbl et~al.(2023)Kerbl, Kopanas, Leimk{\"u}hler, and Drettakis]{kerbl20233d}
Bernhard Kerbl, Georgios Kopanas, Thomas Leimk{\"u}hler, and George Drettakis.
\newblock 3d gaussian splatting for real-time radiance field rendering.
\newblock \emph{ACM Trans. Graph.}, 42\penalty0 (4):\penalty0 139--1, 2023.

\bibitem[Kuang et~al.(2024)Kuang, Cai, He, Xu, Li, Guibas, and Wetzstein]{kuang2024collaborative}
Zhengfei Kuang, Shengqu Cai, Hao He, Yinghao Xu, Hongsheng Li, Leonidas Guibas, and Gordon Wetzstein.
\newblock Collaborative video diffusion: Consistent multi-video generation with camera control.
\newblock \emph{arXiv preprint arXiv:2405.17414}, 2024.

\bibitem[Lepetit et~al.(2009)Lepetit, Moreno-Noguer, and Fua]{lepetit2009ep}
Vincent Lepetit, Francesc Moreno-Noguer, and Pascal Fua.
\newblock Ep n p: An accurate o (n) solution to the p n p problem.
\newblock \emph{International journal of computer vision}, 81:\penalty0 155--166, 2009.

\bibitem[Li et~al.(2022)Li, Slavcheva, Zollhoefer, Green, Lassner, Kim, Schmidt, Lovegrove, Goesele, Newcombe, et~al.]{li2022neural}
Tianye Li, Mira Slavcheva, Michael Zollhoefer, Simon Green, Christoph Lassner, Changil Kim, Tanner Schmidt, Steven Lovegrove, Michael Goesele, Richard Newcombe, et~al.
\newblock Neural 3d video synthesis from multi-view video.
\newblock In \emph{Proceedings of the IEEE/CVF Conference on Computer Vision and Pattern Recognition}, pages 5521--5531, 2022.

\bibitem[Li et~al.(2019)Li, Dekel, Cole, Tucker, Snavely, Liu, and Freeman]{li2019learning}
Zhengqi Li, Tali Dekel, Forrester Cole, Richard Tucker, Noah Snavely, Ce Liu, and William~T Freeman.
\newblock Learning the depths of moving people by watching frozen people.
\newblock In \emph{Proceedings of the IEEE/CVF conference on computer vision and pattern recognition}, pages 4521--4530, 2019.

\bibitem[Li et~al.(2023{\natexlab{a}})Li, Wang, Cole, Tucker, and Snavely]{li2023dynibar}
Zhengqi Li, Qianqian Wang, Forrester Cole, Richard Tucker, and Noah Snavely.
\newblock Dynibar: Neural dynamic image-based rendering.
\newblock In \emph{Proceedings of the IEEE/CVF Conference on Computer Vision and Pattern Recognition}, pages 4273--4284, 2023{\natexlab{a}}.

\bibitem[Li et~al.(2023{\natexlab{b}})Li, Zhu, Han, Hou, Guo, and Cheng]{li2023amt}
Zhen Li, Zuo-Liang Zhu, Ling-Hao Han, Qibin Hou, Chun-Le Guo, and Ming-Ming Cheng.
\newblock Amt: All-pairs multi-field transforms for efficient frame interpolation.
\newblock In \emph{Proceedings of the IEEE/CVF Conference on Computer Vision and Pattern Recognition}, pages 9801--9810, 2023{\natexlab{b}}.

\bibitem[Lindenberger et~al.(2023)Lindenberger, Sarlin, and Pollefeys]{lindenberger2023lightglue}
Philipp Lindenberger, Paul-Edouard Sarlin, and Marc Pollefeys.
\newblock Lightglue: Local feature matching at light speed.
\newblock In \emph{Proceedings of the IEEE/CVF International Conference on Computer Vision}, pages 17627--17638, 2023.

\bibitem[Liu et~al.(2021)Liu, Tucker, Jampani, Makadia, Snavely, and Kanazawa]{liu2021infinite}
Andrew Liu, Richard Tucker, Varun Jampani, Ameesh Makadia, Noah Snavely, and Angjoo Kanazawa.
\newblock Infinite nature: Perpetual view generation of natural scenes from a single image.
\newblock In \emph{Proceedings of the IEEE/CVF International Conference on Computer Vision}, pages 14458--14467, 2021.

\bibitem[Liu et~al.(2024)Liu, Sun, Wang, Wang, Sun, Ye, Zhang, and Duan]{liu2024reconx}
Fangfu Liu, Wenqiang Sun, Hanyang Wang, Yikai Wang, Haowen Sun, Junliang Ye, Jun Zhang, and Yueqi Duan.
\newblock Reconx: Reconstruct any scene from sparse views with video diffusion model.
\newblock \emph{arXiv preprint arXiv:2408.16767}, 2024.

\bibitem[Liu et~al.(2023{\natexlab{a}})Liu, Wu, Van~Hoorick, Tokmakov, Zakharov, and Vondrick]{liu2023zero}
Ruoshi Liu, Rundi Wu, Basile Van~Hoorick, Pavel Tokmakov, Sergey Zakharov, and Carl Vondrick.
\newblock Zero-1-to-3: Zero-shot one image to 3d object.
\newblock In \emph{Proceedings of the IEEE/CVF international conference on computer vision}, pages 9298--9309, 2023{\natexlab{a}}.

\bibitem[Liu et~al.(2023{\natexlab{b}})Liu, Lin, Zeng, Long, Liu, Komura, and Wang]{liu2023syncdreamer}
Yuan Liu, Cheng Lin, Zijiao Zeng, Xiaoxiao Long, Lingjie Liu, Taku Komura, and Wenping Wang.
\newblock Syncdreamer: Generating multiview-consistent images from a single-view image.
\newblock \emph{arXiv preprint arXiv:2309.03453}, 2023{\natexlab{b}}.

\bibitem[Luo et~al.(2020)Luo, Huang, Szeliski, Matzen, and Kopf]{luo2020consistent}
Xuan Luo, Jia-Bin Huang, Richard Szeliski, Kevin Matzen, and Johannes Kopf.
\newblock Consistent video depth estimation.
\newblock \emph{ACM Transactions on Graphics (ToG)}, 39\penalty0 (4):\penalty0 71--1, 2020.

\bibitem[Men et~al.(2024)Men, Yao, Cui, and Bo]{men2024mimo}
Yifang Men, Yuan Yao, Miaomiao Cui, and Liefeng Bo.
\newblock Mimo: Controllable character video synthesis with spatial decomposed modeling.
\newblock \emph{arXiv preprint arXiv:2409.16160}, 2024.

\bibitem[Mildenhall et~al.(2021)Mildenhall, Srinivasan, Tancik, Barron, Ramamoorthi, and Ng]{mildenhall2021nerf}
Ben Mildenhall, Pratul~P Srinivasan, Matthew Tancik, Jonathan~T Barron, Ravi Ramamoorthi, and Ren Ng.
\newblock Nerf: Representing scenes as neural radiance fields for view synthesis.
\newblock \emph{Communications of the ACM}, 65\penalty0 (1):\penalty0 99--106, 2021.

\bibitem[Mou et~al.(2024)Mou, Wang, Xie, Wu, Zhang, Qi, and Shan]{mou2024t2i}
Chong Mou, Xintao Wang, Liangbin Xie, Yanze Wu, Jian Zhang, Zhongang Qi, and Ying Shan.
\newblock T2i-adapter: Learning adapters to dig out more controllable ability for text-to-image diffusion models.
\newblock In \emph{Proceedings of the AAAI Conference on Artificial Intelligence}, pages 4296--4304, 2024.

\bibitem[M{\"u}ller et~al.(2024)M{\"u}ller, Schwarz, R{\"o}ssle, Porzi, Bul{\`o}, Nie{\ss}ner, and Kontschieder]{muller2024multidiff}
Norman M{\"u}ller, Katja Schwarz, Barbara R{\"o}ssle, Lorenzo Porzi, Samuel~Rota Bul{\`o}, Matthias Nie{\ss}ner, and Peter Kontschieder.
\newblock Multidiff: Consistent novel view synthesis from a single image.
\newblock In \emph{Proceedings of the IEEE/CVF Conference on Computer Vision and Pattern Recognition}, pages 10258--10268, 2024.

\bibitem[Niu et~al.(2024)Niu, Cun, Wang, Zhang, Shan, and Zheng]{niu2024mofa}
Muyao Niu, Xiaodong Cun, Xintao Wang, Yong Zhang, Ying Shan, and Yinqiang Zheng.
\newblock Mofa-video: Controllable image animation via generative motion field adaptions in frozen image-to-video diffusion model.
\newblock \emph{arXiv preprint arXiv:2405.20222}, 2024.

\bibitem[Park et~al.(2021)Park, Sinha, Hedman, Barron, Bouaziz, Goldman, Martin-Brualla, and Seitz]{park2021hypernerf}
Keunhong Park, Utkarsh Sinha, Peter Hedman, Jonathan~T Barron, Sofien Bouaziz, Dan~B Goldman, Ricardo Martin-Brualla, and Steven~M Seitz.
\newblock Hypernerf: A higher-dimensional representation for topologically varying neural radiance fields.
\newblock \emph{arXiv preprint arXiv:2106.13228}, 2021.

\bibitem[Piccinelli et~al.(2024)Piccinelli, Yang, Sakaridis, Segu, Li, Van~Gool, and Yu]{piccinelli2024unidepth}
Luigi Piccinelli, Yung-Hsu Yang, Christos Sakaridis, Mattia Segu, Siyuan Li, Luc Van~Gool, and Fisher Yu.
\newblock Unidepth: Universal monocular metric depth estimation.
\newblock In \emph{Proceedings of the IEEE/CVF Conference on Computer Vision and Pattern Recognition}, pages 10106--10116, 2024.

\bibitem[Qi et~al.(2023)Qi, Cun, Zhang, Lei, Wang, Shan, and Chen]{qi2023fatezero}
Chenyang Qi, Xiaodong Cun, Yong Zhang, Chenyang Lei, Xintao Wang, Ying Shan, and Qifeng Chen.
\newblock Fatezero: Fusing attentions for zero-shot text-based video editing.
\newblock In \emph{Proceedings of the IEEE/CVF International Conference on Computer Vision}, pages 15932--15942, 2023.

\bibitem[Radford et~al.(2021)Radford, Kim, Hallacy, Ramesh, Goh, Agarwal, Sastry, Askell, Mishkin, Clark, et~al.]{radford2021learning}
Alec Radford, Jong~Wook Kim, Chris Hallacy, Aditya Ramesh, Gabriel Goh, Sandhini Agarwal, Girish Sastry, Amanda Askell, Pamela Mishkin, Jack Clark, et~al.
\newblock Learning transferable visual models from natural language supervision.
\newblock In \emph{International conference on machine learning}, pages 8748--8763. PMLR, 2021.

\bibitem[Ramasinghe et~al.(2024)Ramasinghe, Shevchenko, Avraham, and Van Den~Hengel]{ramasinghe2024blirf}
Sameera Ramasinghe, Violetta Shevchenko, Gil Avraham, and Anton Van Den~Hengel.
\newblock Blirf: Bandlimited radiance fields for dynamic scene modeling.
\newblock In \emph{Proceedings of the AAAI Conference on Artificial Intelligence}, pages 4641--4649, 2024.

\bibitem[Ranftl et~al.(2020)Ranftl, Lasinger, Hafner, Schindler, and Koltun]{ranftl2020towards}
Ren{\'e} Ranftl, Katrin Lasinger, David Hafner, Konrad Schindler, and Vladlen Koltun.
\newblock Towards robust monocular depth estimation: Mixing datasets for zero-shot cross-dataset transfer.
\newblock \emph{IEEE transactions on pattern analysis and machine intelligence}, 44\penalty0 (3):\penalty0 1623--1637, 2020.

\bibitem[Ranftl et~al.(2021)Ranftl, Bochkovskiy, and Koltun]{ranftl2021vision}
Ren{\'e} Ranftl, Alexey Bochkovskiy, and Vladlen Koltun.
\newblock Vision transformers for dense prediction.
\newblock In \emph{Proceedings of the IEEE/CVF international conference on computer vision}, pages 12179--12188, 2021.

\bibitem[Ravi et~al.(2024)Ravi, Gabeur, Hu, Hu, Ryali, Ma, Khedr, R{\"a}dle, Rolland, Gustafson, et~al.]{ravi2024sam}
Nikhila Ravi, Valentin Gabeur, Yuan-Ting Hu, Ronghang Hu, Chaitanya Ryali, Tengyu Ma, Haitham Khedr, Roman R{\"a}dle, Chloe Rolland, Laura Gustafson, et~al.
\newblock Sam 2: Segment anything in images and videos.
\newblock \emph{arXiv preprint arXiv:2408.00714}, 2024.

\bibitem[Reizenstein et~al.(2021)Reizenstein, Shapovalov, Henzler, Sbordone, Labatut, and Novotny]{reizenstein2021common}
Jeremy Reizenstein, Roman Shapovalov, Philipp Henzler, Luca Sbordone, Patrick Labatut, and David Novotny.
\newblock Common objects in 3d: Large-scale learning and evaluation of real-life 3d category reconstruction.
\newblock In \emph{Proceedings of the IEEE/CVF international conference on computer vision}, pages 10901--10911, 2021.

\bibitem[Rombach et~al.(2021)Rombach, Esser, and Ommer]{rombach2021geometry}
Robin Rombach, Patrick Esser, and Bj{\"o}rn Ommer.
\newblock Geometry-free view synthesis: Transformers and no 3d priors.
\newblock In \emph{Proceedings of the IEEE/CVF International Conference on Computer Vision}, pages 14356--14366, 2021.

\bibitem[Rombach et~al.(2022)Rombach, Blattmann, Lorenz, Esser, and Ommer]{rombach2022high}
Robin Rombach, Andreas Blattmann, Dominik Lorenz, Patrick Esser, and Bj{\"o}rn Ommer.
\newblock High-resolution image synthesis with latent diffusion models.
\newblock In \emph{Proceedings of the IEEE/CVF conference on computer vision and pattern recognition}, pages 10684--10695, 2022.

\bibitem[Sargent et~al.(2023)Sargent, Li, Shah, Herrmann, Yu, Zhang, Chan, Lagun, Fei-Fei, Sun, et~al.]{sargent2023zeronvs}
Kyle Sargent, Zizhang Li, Tanmay Shah, Charles Herrmann, Hong-Xing Yu, Yunzhi Zhang, Eric~Ryan Chan, Dmitry Lagun, Li Fei-Fei, Deqing Sun, et~al.
\newblock Zeronvs: Zero-shot 360-degree view synthesis from a single real image.
\newblock \emph{arXiv preprint arXiv:2310.17994}, 2023.

\bibitem[Sarlin et~al.(2020)Sarlin, DeTone, Malisiewicz, and Rabinovich]{sarlin2020superglue}
Paul-Edouard Sarlin, Daniel DeTone, Tomasz Malisiewicz, and Andrew Rabinovich.
\newblock Superglue: Learning feature matching with graph neural networks.
\newblock In \emph{Proceedings of the IEEE/CVF conference on computer vision and pattern recognition}, pages 4938--4947, 2020.

\bibitem[Schuhmann et~al.(2022)Schuhmann, Beaumont, Vencu, Gordon, Wightman, Cherti, Coombes, Katta, Mullis, Wortsman, et~al.]{schuhmann2022laion}
Christoph Schuhmann, Romain Beaumont, Richard Vencu, Cade Gordon, Ross Wightman, Mehdi Cherti, Theo Coombes, Aarush Katta, Clayton Mullis, Mitchell Wortsman, et~al.
\newblock Laion-5b: An open large-scale dataset for training next generation image-text models.
\newblock \emph{Advances in Neural Information Processing Systems}, 35:\penalty0 25278--25294, 2022.

\bibitem[Seo et~al.(2024)Seo, Fukuda, Shibuya, Narihira, Murata, Hu, Lai, Kim, and Mitsufuji]{seo2024genwarp}
Junyoung Seo, Kazumi Fukuda, Takashi Shibuya, Takuya Narihira, Naoki Murata, Shoukang Hu, Chieh-Hsin Lai, Seungryong Kim, and Yuki Mitsufuji.
\newblock Genwarp: Single image to novel views with semantic-preserving generative warping.
\newblock \emph{arXiv preprint arXiv:2405.17251}, 2024.

\bibitem[Shao et~al.(2024{\natexlab{a}})Shao, Yang, Zhou, Zhang, Shen, Poggi, and Liao]{shao2024learning}
Jiahao Shao, Yuanbo Yang, Hongyu Zhou, Youmin Zhang, Yujun Shen, Matteo Poggi, and Yiyi Liao.
\newblock Learning temporally consistent video depth from video diffusion priors.
\newblock \emph{arXiv preprint arXiv:2406.01493}, 2024{\natexlab{a}}.

\bibitem[Shao et~al.(2024{\natexlab{b}})Shao, Pang, Zheng, Sun, and Liu]{shao2024human4dit}
Ruizhi Shao, Youxin Pang, Zerong Zheng, Jingxiang Sun, and Yebin Liu.
\newblock Human4dit: 360-degree human video generation with 4d diffusion transformer.
\newblock \emph{ACM Transactions on Graphics (TOG)}, 43\penalty0 (6), 2024{\natexlab{b}}.

\bibitem[Shi et~al.(2023)Shi, Wang, Ye, Long, Li, and Yang]{shi2023mvdream}
Yichun Shi, Peng Wang, Jianglong Ye, Mai Long, Kejie Li, and Xiao Yang.
\newblock Mvdream: Multi-view diffusion for 3d generation.
\newblock \emph{arXiv preprint arXiv:2308.16512}, 2023.

\bibitem[Sitzmann et~al.(2021)Sitzmann, Rezchikov, Freeman, Tenenbaum, and Durand]{sitzmann2021light}
Vincent Sitzmann, Semon Rezchikov, Bill Freeman, Josh Tenenbaum, and Fredo Durand.
\newblock Light field networks: Neural scene representations with single-evaluation rendering.
\newblock \emph{Advances in Neural Information Processing Systems}, 34:\penalty0 19313--19325, 2021.

\bibitem[Song et~al.(2020)Song, Meng, and Ermon]{song2020denoising}
Jiaming Song, Chenlin Meng, and Stefano Ermon.
\newblock Denoising diffusion implicit models.
\newblock \emph{arXiv preprint arXiv:2010.02502}, 2020.

\bibitem[Stan et~al.(2023)Stan, Wofk, Aflalo, Tseng, Cai, Paulitsch, and Lal]{stan2023ldm3d}
Gabriela Ben~Melech Stan, Diana Wofk, Estelle Aflalo, Shao-Yen Tseng, Zhipeng Cai, Michael Paulitsch, and Vasudev Lal.
\newblock Ldm3d-vr: Latent diffusion model for 3d vr.
\newblock \emph{arXiv preprint arXiv:2311.03226}, 2023.

\bibitem[Teed and Deng(2020)]{teed2020raft}
Zachary Teed and Jia Deng.
\newblock Raft: Recurrent all-pairs field transforms for optical flow.
\newblock In \emph{Computer Vision--ECCV 2020: 16th European Conference, Glasgow, UK, August 23--28, 2020, Proceedings, Part II 16}, pages 402--419. Springer, 2020.

\bibitem[Teng et~al.(2023)Teng, Xie, Wu, Han, Li, and Liu]{teng2023drag}
Yao Teng, Enze Xie, Yue Wu, Haoyu Han, Zhenguo Li, and Xihui Liu.
\newblock Drag-a-video: Non-rigid video editing with point-based interaction.
\newblock \emph{arXiv preprint arXiv:2312.02936}, 2023.

\bibitem[Truong et~al.(2023)Truong, Rakotosaona, Manhardt, and Tombari]{truong2023sparf}
Prune Truong, Marie-Julie Rakotosaona, Fabian Manhardt, and Federico Tombari.
\newblock Sparf: Neural radiance fields from sparse and noisy poses.
\newblock In \emph{Proceedings of the IEEE/CVF Conference on Computer Vision and Pattern Recognition}, pages 4190--4200, 2023.

\bibitem[Tung et~al.(2025)Tung, Chou, Cai, Yang, Zhang, Wetzstein, Hariharan, and Snavely]{tung2025megascenes}
Joseph Tung, Gene Chou, Ruojin Cai, Guandao Yang, Kai Zhang, Gordon Wetzstein, Bharath Hariharan, and Noah Snavely.
\newblock Megascenes: Scene-level view synthesis at scale.
\newblock In \emph{European Conference on Computer Vision}, pages 197--214. Springer, 2025.

\bibitem[Van~Hoorick et~al.(2024)Van~Hoorick, Wu, Ozguroglu, Sargent, Liu, Tokmakov, Dave, Zheng, and Vondrick]{van2024generative}
Basile Van~Hoorick, Rundi Wu, Ege Ozguroglu, Kyle Sargent, Ruoshi Liu, Pavel Tokmakov, Achal Dave, Changxi Zheng, and Carl Vondrick.
\newblock Generative camera dolly: Extreme monocular dynamic novel view synthesis.
\newblock \emph{arXiv preprint arXiv:2405.14868}, 2024.

\bibitem[Wang et~al.(2024{\natexlab{a}})Wang, Zhuang, Siarohin, Cao, Qian, Lee, and Tulyakov]{wang2024diffusion}
Chaoyang Wang, Peiye Zhuang, Aliaksandr Siarohin, Junli Cao, Guocheng Qian, Hsin-Ying Lee, and Sergey Tulyakov.
\newblock Diffusion priors for dynamic view synthesis from monocular videos.
\newblock \emph{arXiv preprint arXiv:2401.05583}, 2024{\natexlab{a}}.

\bibitem[Wang and Shi(2023)]{wang2023imagedream}
Peng Wang and Yichun Shi.
\newblock Imagedream: Image-prompt multi-view diffusion for 3d generation.
\newblock \emph{arXiv preprint arXiv:2312.02201}, 2023.

\bibitem[Wang et~al.(2024{\natexlab{b}})Wang, Ye, Gao, Austin, Li, and Kanazawa]{wang2024shape}
Qianqian Wang, Vickie Ye, Hang Gao, Jake Austin, Zhengqi Li, and Angjoo Kanazawa.
\newblock Shape of motion: 4d reconstruction from a single video.
\newblock \emph{arXiv preprint arXiv:2407.13764}, 2024{\natexlab{b}}.

\bibitem[Wang et~al.(2024{\natexlab{c}})Wang, Leroy, Cabon, Chidlovskii, and Revaud]{wang2024dust3r}
Shuzhe Wang, Vincent Leroy, Yohann Cabon, Boris Chidlovskii, and Jerome Revaud.
\newblock Dust3r: Geometric 3d vision made easy.
\newblock In \emph{Proceedings of the IEEE/CVF Conference on Computer Vision and Pattern Recognition}, pages 20697--20709, 2024{\natexlab{c}}.

\bibitem[Wang et~al.(2020)Wang, Yu, Zhao, Zhu, Qin, Zhou, Zhou, and Lei]{wang2020deep}
Zezheng Wang, Zitong Yu, Chenxu Zhao, Xiangyu Zhu, Yunxiao Qin, Qiusheng Zhou, Feng Zhou, and Zhen Lei.
\newblock Deep spatial gradient and temporal depth learning for face anti-spoofing.
\newblock In \emph{Proceedings of the IEEE/CVF conference on computer vision and pattern recognition}, pages 5042--5051, 2020.

\bibitem[Wang et~al.(2024{\natexlab{d}})Wang, Yuan, Wang, Li, Chen, Xia, Luo, and Shan]{wang2024motionctrl}
Zhouxia Wang, Ziyang Yuan, Xintao Wang, Yaowei Li, Tianshui Chen, Menghan Xia, Ping Luo, and Ying Shan.
\newblock Motionctrl: A unified and flexible motion controller for video generation.
\newblock In \emph{ACM SIGGRAPH 2024 Conference Papers}, pages 1--11, 2024{\natexlab{d}}.

\bibitem[Watson et~al.(2024)Watson, Saxena, Li, Tagliasacchi, and Fleet]{watson2024controlling}
Daniel Watson, Saurabh Saxena, Lala Li, Andrea Tagliasacchi, and David~J Fleet.
\newblock Controlling space and time with diffusion models.
\newblock \emph{arXiv preprint arXiv:2407.07860}, 2024.

\bibitem[Wu et~al.(2024{\natexlab{a}})Wu, Yi, Fang, Xie, Zhang, Wei, Liu, Tian, and Wang]{wu20244d}
Guanjun Wu, Taoran Yi, Jiemin Fang, Lingxi Xie, Xiaopeng Zhang, Wei Wei, Wenyu Liu, Qi Tian, and Xinggang Wang.
\newblock 4d gaussian splatting for real-time dynamic scene rendering.
\newblock In \emph{Proceedings of the IEEE/CVF Conference on Computer Vision and Pattern Recognition}, pages 20310--20320, 2024{\natexlab{a}}.

\bibitem[Wu et~al.(2024{\natexlab{b}})Wu, Gao, Poole, Trevithick, Zheng, Barron, and Holynski]{wu2024cat4d}
Rundi Wu, Ruiqi Gao, Ben Poole, Alex Trevithick, Changxi Zheng, Jonathan~T Barron, and Aleksander Holynski.
\newblock Cat4d: Create anything in 4d with multi-view video diffusion models.
\newblock \emph{arXiv preprint arXiv:2411.18613}, 2024{\natexlab{b}}.

\bibitem[Xu et~al.(2024)Xu, Nie, Liu, Liu, Kautz, Wang, and Vahdat]{xu2024camco}
Dejia Xu, Weili Nie, Chao Liu, Sifei Liu, Jan Kautz, Zhangyang Wang, and Arash Vahdat.
\newblock Camco: Camera-controllable 3d-consistent image-to-video generation.
\newblock \emph{arXiv preprint arXiv:2406.02509}, 2024.

\bibitem[Yang et~al.(2024{\natexlab{a}})Yang, Huang, Yin, Shen, Liu, He, Lin, Ouyang, and He]{yang2024depth}
Honghui Yang, Di Huang, Wei Yin, Chunhua Shen, Haifeng Liu, Xiaofei He, Binbin Lin, Wanli Ouyang, and Tong He.
\newblock Depth any video with scalable synthetic data.
\newblock \emph{arXiv preprint arXiv:2410.10815}, 2024{\natexlab{a}}.

\bibitem[Yang et~al.(2024{\natexlab{b}})Yang, Kang, Huang, Xu, Feng, and Zhao]{yang2024depthany}
Lihe Yang, Bingyi Kang, Zilong Huang, Xiaogang Xu, Jiashi Feng, and Hengshuang Zhao.
\newblock Depth anything: Unleashing the power of large-scale unlabeled data.
\newblock In \emph{Proceedings of the IEEE/CVF Conference on Computer Vision and Pattern Recognition}, pages 10371--10381, 2024{\natexlab{b}}.

\bibitem[Yang et~al.(2024{\natexlab{c}})Yang, Kang, Huang, Zhao, Xu, Feng, and Zhao]{yang2024depthany2}
Lihe Yang, Bingyi Kang, Zilong Huang, Zhen Zhao, Xiaogang Xu, Jiashi Feng, and Hengshuang Zhao.
\newblock Depth anything v2.
\newblock \emph{arXiv preprint arXiv:2406.09414}, 2024{\natexlab{c}}.

\bibitem[Yang et~al.(2023)Yang, Yang, Pan, and Zhang]{yang2023real}
Zeyu Yang, Hongye Yang, Zijie Pan, and Li Zhang.
\newblock Real-time photorealistic dynamic scene representation and rendering with 4d gaussian splatting.
\newblock \emph{arXiv preprint arXiv:2310.10642}, 2023.

\bibitem[Yang et~al.(2024{\natexlab{d}})Yang, Gao, Zhou, Jiao, Zhang, and Jin]{yang2024deformable}
Ziyi Yang, Xinyu Gao, Wen Zhou, Shaohui Jiao, Yuqing Zhang, and Xiaogang Jin.
\newblock Deformable 3d gaussians for high-fidelity monocular dynamic scene reconstruction.
\newblock In \emph{Proceedings of the IEEE/CVF Conference on Computer Vision and Pattern Recognition}, pages 20331--20341, 2024{\natexlab{d}}.

\bibitem[Yang et~al.(2024{\natexlab{e}})Yang, Teng, Zheng, Ding, Huang, Xu, Yang, Hong, Zhang, Feng, et~al.]{yang2024cogvideox}
Zhuoyi Yang, Jiayan Teng, Wendi Zheng, Ming Ding, Shiyu Huang, Jiazheng Xu, Yuanming Yang, Wenyi Hong, Xiaohan Zhang, Guanyu Feng, et~al.
\newblock Cogvideox: Text-to-video diffusion models with an expert transformer.
\newblock \emph{arXiv preprint arXiv:2408.06072}, 2024{\natexlab{e}}.

\bibitem[Yu et~al.(2020)Yu, Zhao, Wang, Qin, Su, Li, Zhou, and Zhao]{yu2020searching}
Zitong Yu, Chenxu Zhao, Zezheng Wang, Yunxiao Qin, Zhuo Su, Xiaobai Li, Feng Zhou, and Guoying Zhao.
\newblock Searching central difference convolutional networks for face anti-spoofing.
\newblock In \emph{Proceedings of the IEEE/CVF conference on computer vision and pattern recognition}, pages 5295--5305, 2020.

\bibitem[Zhang et~al.(2024{\natexlab{a}})Zhang, Paiss, Zada, Karnad, Jacobs, Pritch, Mosseri, Shou, Wadhwa, and Ruiz]{zhang2024recapture}
David~Junhao Zhang, Roni Paiss, Shiran Zada, Nikhil Karnad, David~E Jacobs, Yael Pritch, Inbar Mosseri, Mike~Zheng Shou, Neal Wadhwa, and Nataniel Ruiz.
\newblock Recapture: Generative video camera controls for user-provided videos using masked video fine-tuning.
\newblock \emph{arXiv preprint arXiv:2411.05003}, 2024{\natexlab{a}}.

\bibitem[Zhang et~al.(2021{\natexlab{a}})Zhang, Liu, Ye, Zhao, Zhang, Wu, Zhang, Yu, and Xu]{zhang2021editable}
Jiakai Zhang, Xinhang Liu, Xinyi Ye, Fuqiang Zhao, Yanshun Zhang, Minye Wu, Yingliang Zhang, Jingyi Yu, and Lan Xu.
\newblock Editable free-viewpoint video using a layered neural representation.
\newblock 2021{\natexlab{a}}.

\bibitem[Zhang et~al.(2024{\natexlab{b}})Zhang, Herrmann, Hur, Jampani, Darrell, Cole, Sun, and Yang]{zhang2024monst3r}
Junyi Zhang, Charles Herrmann, Junhwa Hur, Varun Jampani, Trevor Darrell, Forrester Cole, Deqing Sun, and Ming-Hsuan Yang.
\newblock Monst3r: A simple approach for estimating geometry in the presence of motion.
\newblock \emph{arXiv preprint arXiv:2410.03825}, 2024{\natexlab{b}}.

\bibitem[Zhang et~al.(2021{\natexlab{b}})Zhang, Cole, Tucker, Freeman, and Dekel]{zhang2021consistent}
Zhoutong Zhang, Forrester Cole, Richard Tucker, William~T Freeman, and Tali Dekel.
\newblock Consistent depth of moving objects in video.
\newblock \emph{ACM Transactions on Graphics (ToG)}, 40\penalty0 (4):\penalty0 1--12, 2021{\natexlab{b}}.

\bibitem[Zhang et~al.(2024{\natexlab{c}})Zhang, Liao, Li, Qin, and Wang]{zhang2024tora}
Zhenghao Zhang, Junchao Liao, Menghao Li, Long Qin, and Weizhi Wang.
\newblock Tora: Trajectory-oriented diffusion transformer for video generation.
\newblock \emph{arXiv preprint arXiv:2407.21705}, 2024{\natexlab{c}}.

\bibitem[Zhao et~al.(2024)Zhao, Colburn, Ma, Bautista, Susskind, and Schwing]{zhao2024pseudo}
Xiaoming Zhao, R~Alex Colburn, Fangchang Ma, Miguel~{\'A}ngel Bautista, Joshua~M Susskind, and Alex Schwing.
\newblock Pseudo-generalized dynamic view synthesis from a video.
\newblock In \emph{The Twelfth International Conference on Learning Representations}, 2024.

\bibitem[Zhou et~al.(2018)Zhou, Tucker, Flynn, Fyffe, and Snavely]{zhou2018stereo}
Tinghui Zhou, Richard Tucker, John Flynn, Graham Fyffe, and Noah Snavely.
\newblock Stereo magnification: Learning view synthesis using multiplane images.
\newblock \emph{arXiv preprint arXiv:1805.09817}, 2018.

\bibitem[Zi et~al.(2024)Zi, Zhao, Qi, Wang, Shi, Chen, Liang, Wong, and Zhang]{zi2024cococo}
Bojia Zi, Shihao Zhao, Xianbiao Qi, Jianan Wang, Yukai Shi, Qianyu Chen, Bin Liang, Kam-Fai Wong, and Lei Zhang.
\newblock Cococo: Improving text-guided video inpainting for better consistency, controllability and compatibility.
\newblock \emph{arXiv preprint arXiv:2403.12035}, 2024.

\end{thebibliography}
}

\clearpage

\clearpage
\maketitlesupplementary
\appendix

In this appendix, we provide:

\begin{itemize}
\item \textbf{Additional Experimental Results:} We present more qualitative examples that showcase the robustness and effectiveness of our method in various scenarios. We also provide video results in Supplementary Materials.
\item \textbf{More Details:} We detail our architecture and training procedures and discuss how our approach can incorporate various geometry estimation models, including monocular video depth estimation models. 
\item \textbf{Additional Discussions:} We discuss the challenges in multi-view aggregation for dynamic novel view synthesis and explain how our method effectively handles these without explicit dynamic region masking. 
\item \textbf{Additional Implementation Details:} We provide specifics about our training setups. 
\item \textbf{Details on Experiments:} We elaborate on our user study, the evaluation metrics used, and the experimental setups for baselines.
\end{itemize}

\section{Additional Experimental Results}
\label{sec:appendix_additional_exps}     

\begin{figure*}[t]
    \centering
    \includegraphics[width=\textwidth]{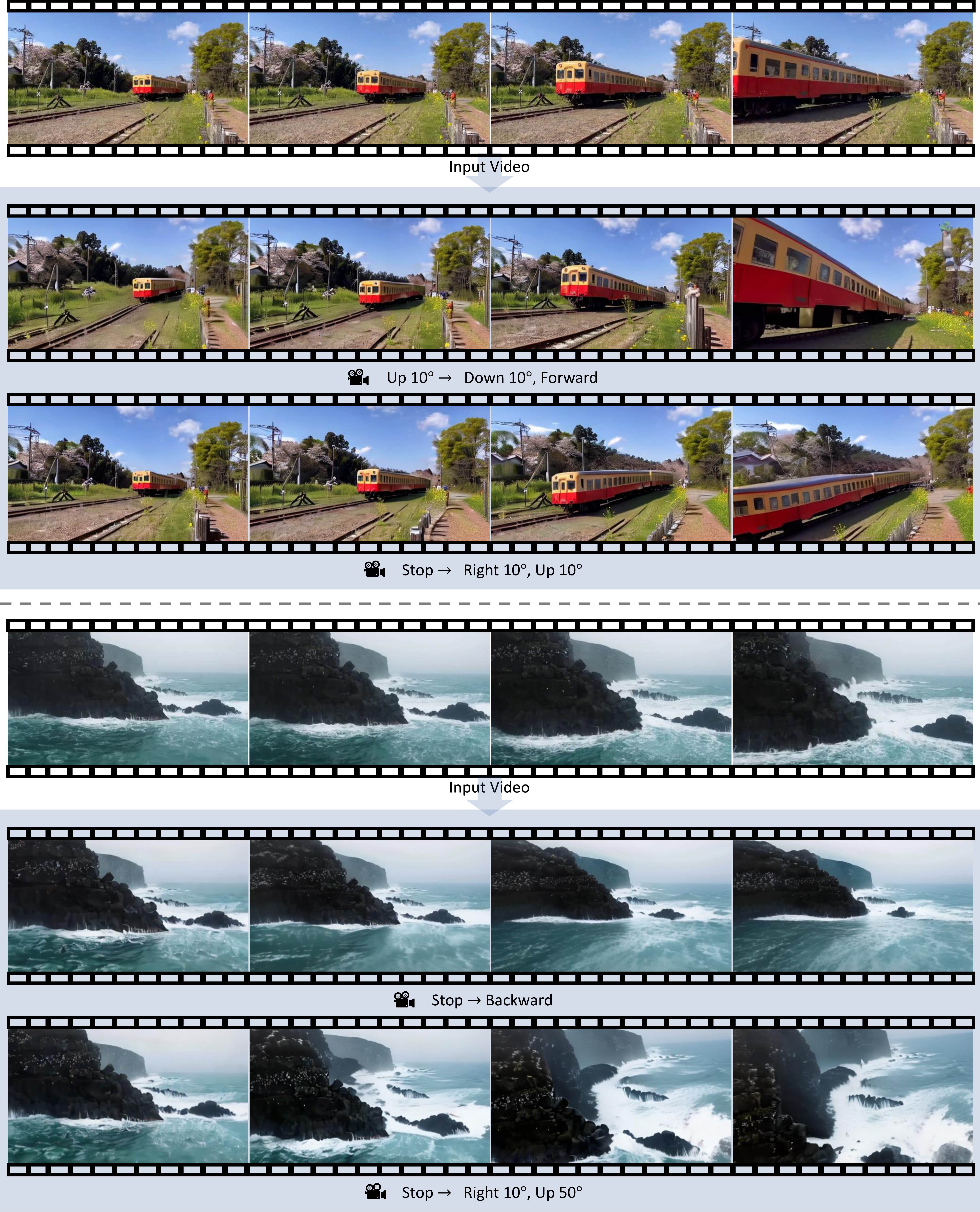} 
    \vspace{-10pt}
    \caption{\textbf{More qualitative results.} We rendered the video using two different camera trajectories for each input video.}
    \vspace{-5pt}
    
    \label{fig:appendix_qual1}
\end{figure*}

\begin{figure*}[t]
    \centering
    \includegraphics[width=\textwidth]{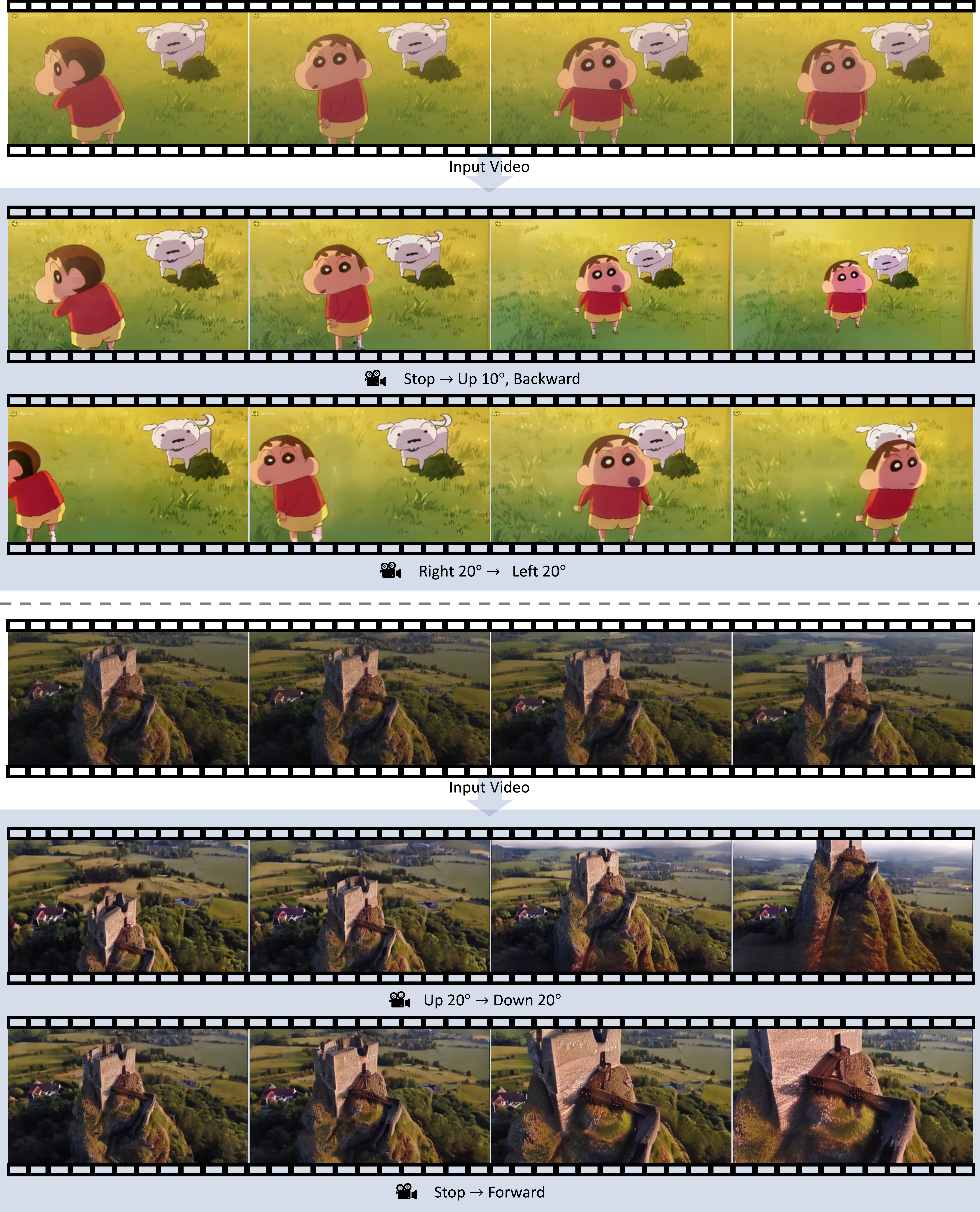} 
    \vspace{-10pt}
    \caption{\textbf{More qualitative results.} We rendered the video using two different camera trajectories for each input video.}
    \vspace{-5pt}
    
    \label{fig:appendix_qual2}
\end{figure*}

\paragraph{Additional qualitative results.}
Fig.~\ref{fig:appendix_qual1} and Fig.~\ref{fig:appendix_qual2} show additional qualitative results of our method, showcasing its performance in different scenarios. These examples highlight the robustness and effectiveness of our approach.

\begin{figure*}[t]
    \centering
    \includegraphics[width=\textwidth]{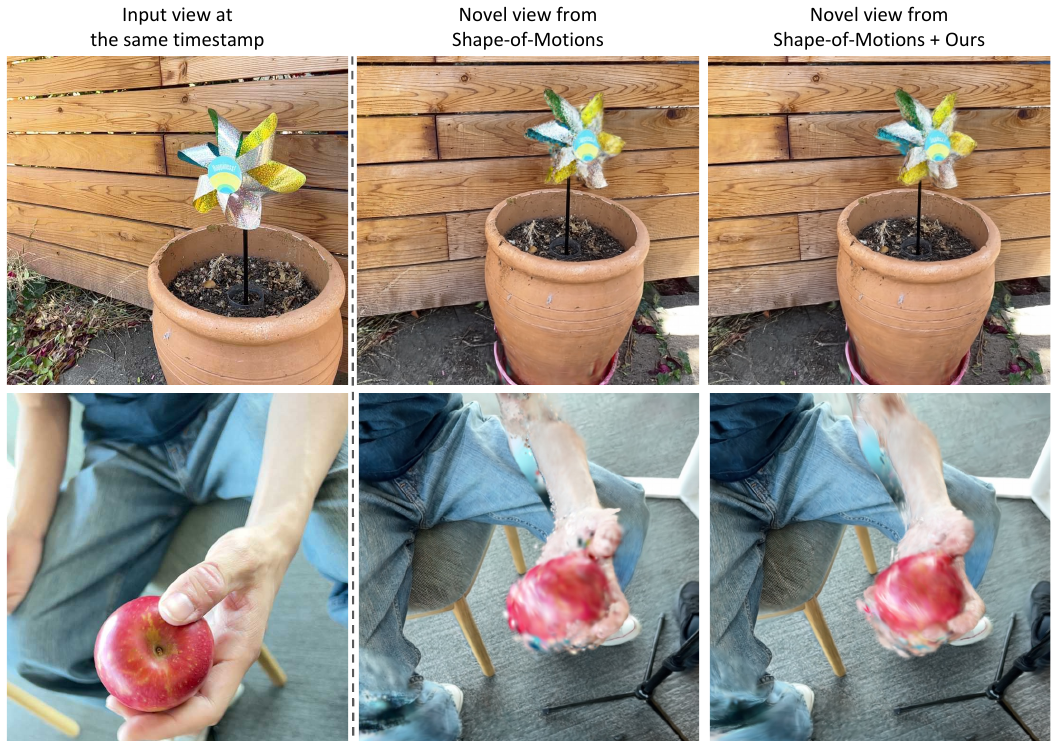} 
    \vspace{-10pt}
    \caption{\textbf{Qualitative results of per-scene reconstruction with our method on DyCheck~\cite{gao2022monocular}.} As an application, the outputs generated by our method can be integrated into optimization-based frameworks for per-scene reconstruction. This helps reduce artifacts in challenging novel views of dynamic 3D representations.}
    \vspace{-5pt}
    
    \label{fig:appendix_dycheck}
\end{figure*}

\paragrapht{Qualitative results on per-scene reconstruction.}   
Our approach can be seamlessly integrated into optimization-based per-scene reconstruction pipelines by using our model’s generated renderings as additional supervision. As shown in Fig.~\ref{fig:appendix_dycheck}, incorporating our generated outputs into the Shape-of-Motion~\cite{wang2024shape} framework improves the performance over the baseline. To generate novel views during training, we periodically generate 24 random camera transformations every 50 epochs. These camera poses enable us to render novel views based on the source views produced by Shape-of-Motion. From these 24 generated novel views, we select two views per iteration as additional training data following~\cite{haque2023instruct}, further enhancing the model's performance.  

\label{sec:appendix_exps}

\section{More Details}
\label{sec:appendix_geometry}

\paragraph{Implementation Details}
\label{sec:appendix_impl}
We trained our model on video and image data at two resolutions: $512 \times 384$ and $512 \times 288$. Specifically, for multi-view image pairs, we trained the model for a total of 344k steps with a batch size of 192. For video data, we fine-tuned the model for 30k steps with a batch size of 16. We employed the 8-bit Adam optimizer with a learning rate of $1 \times 10^{-5}$, and other hyperparameter settings were identical to those used in the training of Stable Diffusion~\cite{rombach2022high} 1.5.
During video training, we randomly removed the video encoder with a probability of 0.3 at each training step to prevent overfitting. In inference, we utilized the DDIM~\cite{song2020denoising} sampler with 20 denoising steps and conditioned the model using classifier-free guidance~\cite{ho2022classifier}.

\begin{figure*}[t]
    \centering
    \includegraphics[width=\textwidth]{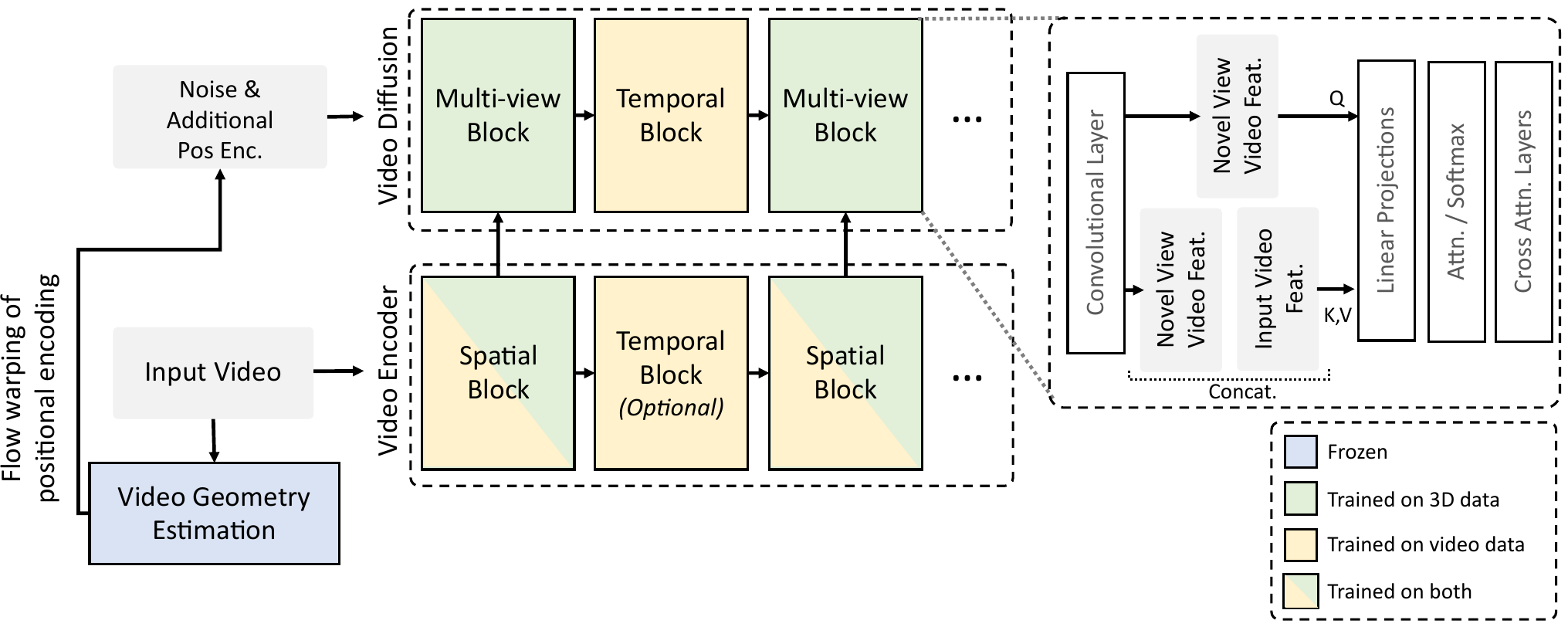} 
    \vspace{-10pt}
    \caption{\textbf{Architecture.} We provide a more detailed illustration of our architecture. In the video diffusion model, which consists of spatial and temporal blocks, we enhance the spatial block by integrating tokens provided by the video encoder, thereby transforming it into a multi-view block. The tokens encoded by the video encoder are concatenated as key and value inputs within the self-attention layers of the multi-view block in the video diffusion model.}
    \vspace{-5pt}
    
    \label{fig:appendix_diagram}
\end{figure*}

\paragraph{Architecture.}
Our method adopts AnimateDiff~\cite{guo2023animatediff}, a video generation model based on Stable Diffusion~\cite{rombach2022high} 1.5, and integrates a video encoder following the ReferenceNet~\cite{hu2024animate} architecture. Specifically, the weights of the video encoder are initialized by the weights of the pre-trained diffusion model. The video encoder extracts the key and value tokens from each self-attention layer block and concatenates these tokens into the corresponding self-attention maps of the video diffusion U-Net.
To condition the video generation model on the 2D flow $f_\mathrm{rel}$ between the input video and the target video, we adopt a positional encoding warping technique proposed in GenWarp~\cite{seo2024genwarp}. We derive both the sinusoidal coordinate map and the flow-warped coordinate map of $f_\mathrm{rel}$. Each of these maps is then processed by a pose encoder network~\cite{mou2024t2i} after the first convolution layer of the video encoder and the diffusion U-Net, where they are subsequently added. A detailed illustration of our model architecture is provided in Fig.~\ref{fig:appendix_diagram}.

\paragraph{Training.}
We use both video data~\cite{jafarian2021learning, bain2021frozen} and multi-view image data~\cite{zhou2018stereo, li2019learning, tung2025megascenes, dai2017scannet} for training. As illustrated in Fig.~\ref{fig:appendix_diagram}, we fine-tune the temporal blocks (colored in yellow) within the diffusion U-Net using only the video data. Similarly, we fine-tune the multi-view blocks (colored in green), using only multi-view image data. Note that multi-view blocks are the spatial blocks of the diffusion U-Net integrated with the feature tokens from the video encoder.

\begin{itemize}
    \item \textbf{Temporal Blocks:} When training the temporal blocks with video data, we freeze the multi-view blocks. In addition, we also freeze the video encoder with a pre-defined probability of $0.5$. This is because when training with video data, the input to the video encoder is identical to what the video diffusion model aims to generate, which can lead to overfitting. When the video encoder is frozen, it becomes equivalent to training the video diffusion model conditioned only on the CLIP image feature of the frames.

    \item \textbf{Multi-view Blocks:} When training the multi-view blocks with multi-view images, we freeze the temporal blocks. As images do not have any temporal length, the video encoder takes the source image and simply treats it as a video with $T=1$ frame and is trained along with the multi-view blocks. To obtain $f_\mathrm{rel}$ between the input and target video, we obtain the depth map of each source view and the corresponding relative camera pose between the source view and the target view prior to training. In datasets~\cite{zhou2018stereo, li2019learning, tung2025megascenes} where dense depth maps are not available, we utilize DUSt3R~\cite{wang2024dust3r} to compute the depth map of the source view and then recompute the corresponding relative camera pose with PnP-RANSAC~\cite{fischler1981random, hartley2003multiple, lepetit2009ep}.
\end{itemize}

\paragraph{Ours with monocular video depth estimation models.}
In our framework, we employ MonST3R~\cite{zhang2024monst3r} as the temporal geometry estimator $g$, enabling the simultaneous prediction of consistent depth maps and camera poses. However, our framework is \textit{versatile} as it can integrate any geometry estimation model for $g$, including monocular or video depth estimation models~\cite{piccinelli2024unidepth, yang2024depthany, yang2024depthany2, hu2024depthcrafter, yang2024depth, shao2024learning}. When using a depth model as $g$, the geometry $\mathcal{G}\in \mathbb{R}^{T\times H\times W\times 3}$ of the input video can be achieved by lifting 2D pixels into 3D space as follows:
\begin{equation}
    \mathcal G = h(K^{-1}D_t), \quad \forall t \in [1, T],
\end{equation}
where $D_t$ represents the predicted depth of the $t$-th frame, and $h: (x,y,z)\rightarrow (x,y,z,1)$ denotes the homogeneous mapping function. 

As depth models cannot infer relative camera poses between images, if the input video is captured from a moving camera, we additionally adopt a 2D correspondence network~\cite{lindenberger2023lightglue, edstedt2024roma, sarlin2020superglue}. This enables us to derive the relative camera poses between frames using PnP-RANSAC~\cite{fischler1981random, hartley2003multiple, lepetit2009ep} from the estimated depth and correspondences. When the input video is captured from a stationary camera, we simply define the camera pose as the identity matrix to obtain the geometry $\mathcal{G}$. For both cases, given the relative camera transformations $C_{\text{rel}}$ for the desired trajectory, our framework can synthesize a novel trajectory video using the same formulation in Eq.~\ref{eq:sample}.

\section{Additional Discussions}
\label{sec:appendix_discussion}
\paragraph{Multi-view aggregation for dynamic novel view synthesis.}
For dynamic novel view synthesis, it is crucial to leverage information from other frames when generating a novel view at an arbitrary video frame. For example, consider an input video moving from left to right, capturing the scene of a living room. At $t=0$, the input video shows a chair, and at $t=5$, it reveals a sofa to the right of the chair. If we aim to synthesize a video from a novel viewpoint that looks from the right, the first frame of the synthesized video should include both the chair and the sofa.

\begin{figure*}[t]
    \centering
    \includegraphics[width=\textwidth]{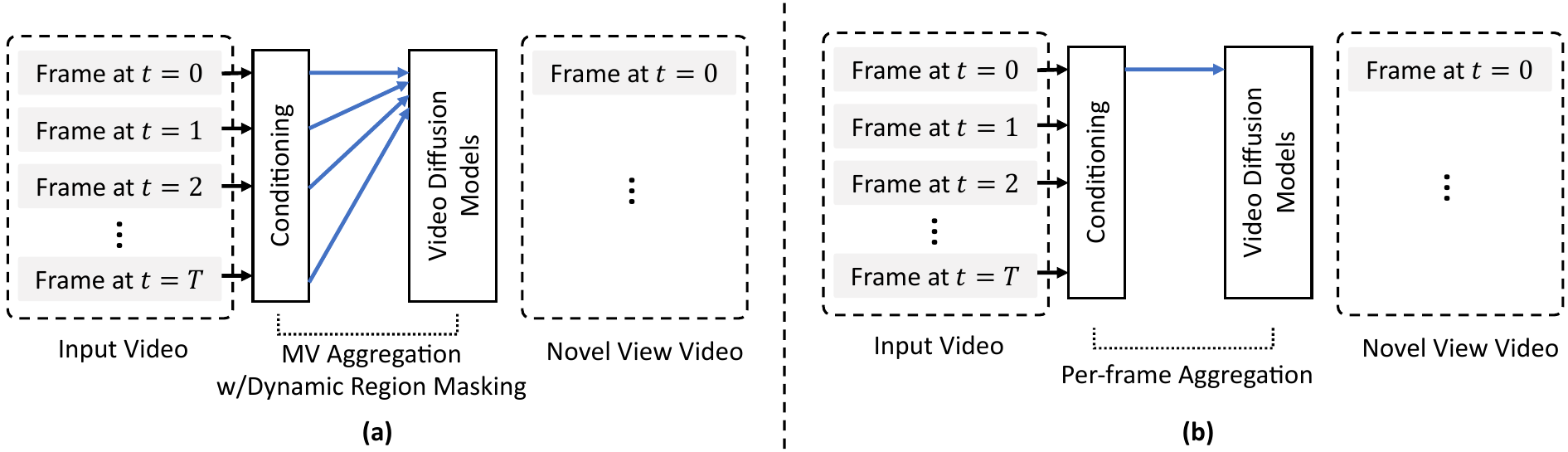} 
    \vspace{-10pt}
    \caption{\textbf{Two approaches to condition the novel view video generation on an input view video.} (a) With estimated mask sequence for dynamic regions, generating each frame of novel view video is conditioned on all the frames following the dynamic region masking. (b) Generating each frame of novel view video is conditioned on a corresponding single frame of input video. The subsequent video diffusion model implicitly achieves the multi-view aggregation during its video generation process.}
    \vspace{-5pt}
    
    \label{fig:appendix_mvs}
\end{figure*}

To achieve this, it is necessary to distinguish between dynamic and static regions, as only the static regions from other frames should be aggregated. MonST3R~\cite{zhang2024monst3r} predicts dynamic regions by thresholding the difference between optical flow and scene flow, which is obtained through the predicted depth and relative camera pose. As the estimated dynamic regions are often noisy, it is further refined with segmentation models such as SAM2~\cite{ravi2024sam}. With the estimated static and dynamic regions, a natural approach to incorporate this into our method is to condition the generation of the $t$-th frame of the novel view video not only on the $t$-th frame of the input video but also on the static regions from other frames, as depicted in Fig.~\ref{fig:appendix_mvs}-(a).

\begin{figure*}[t]
    \centering
    \includegraphics[width=0.85\textwidth]{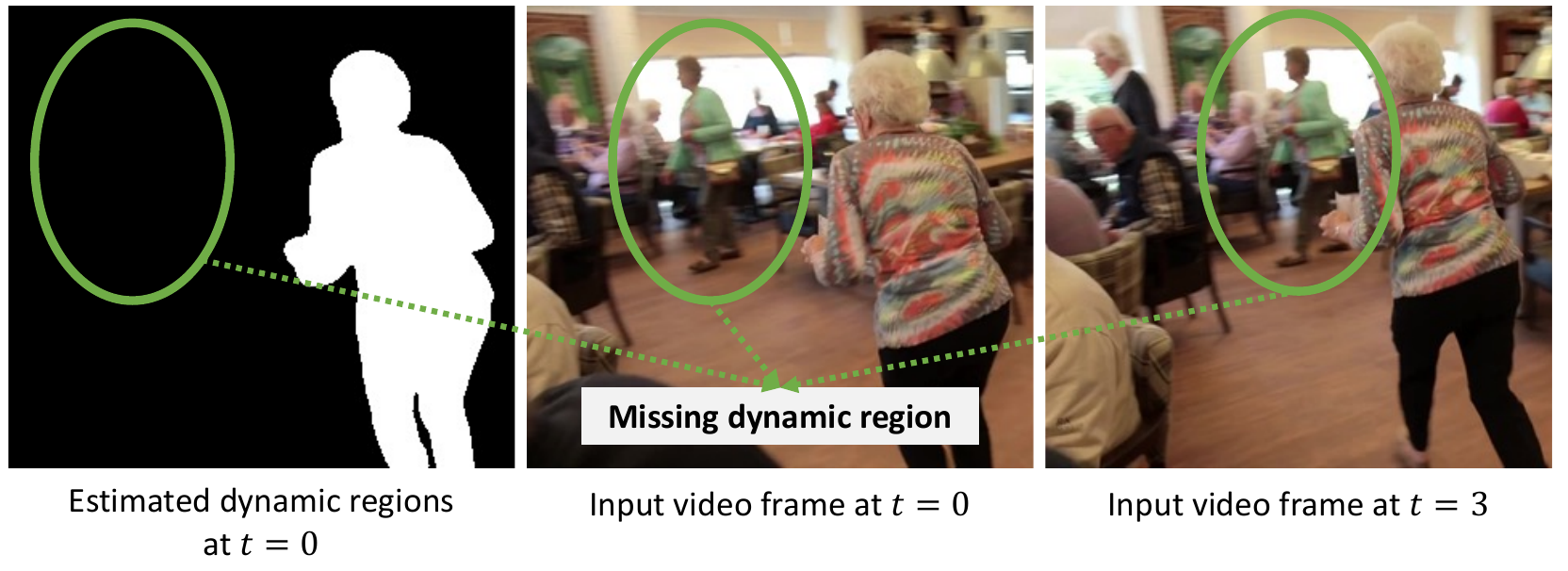} 
    \vspace{-10pt}
    \caption{\textbf{Failure case of dynamic region masking.} 
The dynamic region estimation used in MonST3R~\cite{zhang2024monst3r} effectively captures major dynamic objects but fails to detect more subtle dynamic elements. In the video above, the movements of other people besides the person running on the right are not detected, causing all areas except for the main character to be considered static.
    }
    \vspace{-5pt}
    
    \label{fig:appendix_mvs_eg}
\end{figure*}

However, as shown in Fig.~\ref{fig:appendix_mvs_eg}, these heuristic methods only capture moving objects that are easily distinguishable in the video and fail to consider subtle dynamic elements. Due to this limitation, such methods are applicable to a relatively narrow range of videos. This led us to adopt an approach that leverages the video prior inherent in the video diffusion model, rather than explicitly considering dynamic and static regions, as illustrated in Fig.~\ref{fig:appendix_mvs}-(b). Specifically, when synthesizing the $t$-th frame, we directly provide only the $t$-th frame of the input view and its corresponding 2D flow information. The temporal aggregation within the video generation model implicitly gathers information from other frames. This enables our framework to successfully aggregate multi-view information without being affected by the errors in the predicted dynamic regions.

\section{Details on Experiments}
\label{sec:neu3dstnerf}
\paragraph{Experiments on Neu3D and ST-NeRF dataset.} When evaluating the novel view synthesis quality on Neu3D~\cite{li2022neural} and ST-NeRF~\cite{zhang2021editable} shown in Table~\ref{tab:main_quan}, we designate one camera viewpoint as the source viewpoint and all other camera viewpoints as target viewpoints. Specifically, for Neu3D, we designate the `cam0' sequence as the source video and use all other videos as targets for all scenes. For the ST-NeRF dataset, we choose the middle camera as the source viewpoint, which is `cam7' of each scene. Similar to the Neu3D dataset, we use all other videos as targets for all scenes. We report the average metrics of all scenes for both datasets.

\begin{figure}[h]
    \centering
    \includegraphics[width=0.49\textwidth]{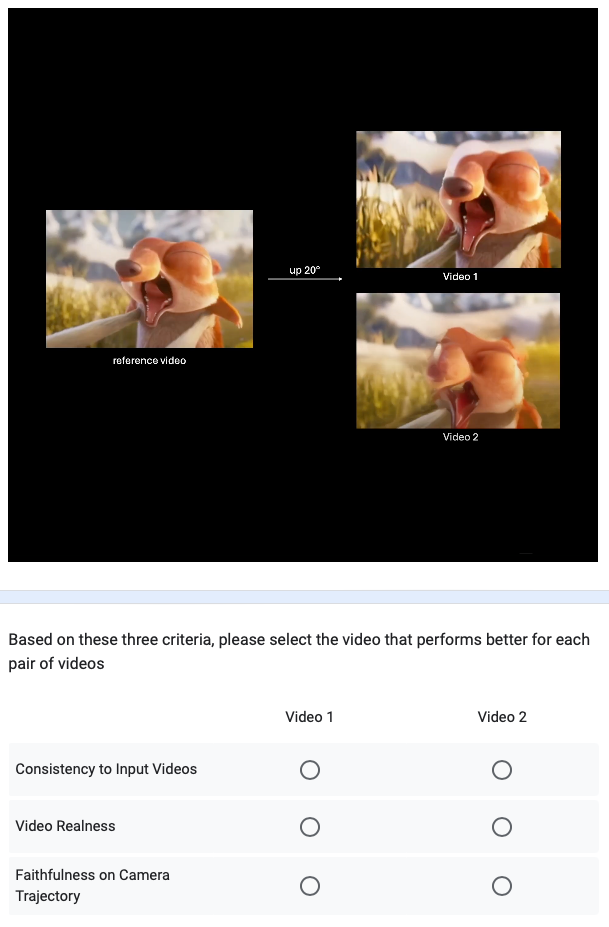} 
    \vspace{-10pt}
    \caption{\textbf{Example of the screen shown to participants.} The order of the results for our method and the baseline methods in each question was randomly shuffled and thoroughly anonymized.}
    \vspace{-5pt}
    
    \label{fig:appendix_userstudy}
\end{figure}

\label{sec:appendix_userstudy}
\paragraph{User study.}
To assess the performance of our model, we conducted a user study, where the results are shown in Fig.~\ref{fig:userstudy}. In the user study, we presented videos generated by our method and Generative Camera Dolly~\cite{van2024generative}, using the same input videos and camera trajectories. A total of 16 input videos and corresponding camera trajectories were randomly sampled for evaluation. Participants were tasked with comparing the generated results based on the following criteria:
\begin{itemize}
    \item \textbf{Consistency to Input Videos:} How closely the generated video resembles the reference video.
    \item \textbf{Video Realness:} How realistic the generated video appears.
    \item \textbf{Faithfulness on Camera Trajectory:} How accurately the generated video follows the given camera movement.
\end{itemize}
To ensure an unbiased evaluation, all videos used in the study were uncurated, and the methods generating each video were not disclosed to the participants. The presentation order of the results was randomized. An example of the interface shown to participants is provided in Fig.~\ref{fig:appendix_userstudy}.

\paragraph{VBench~\cite{huang2024vbench}.}
To quantify and evaluate the quality of the re-synthesized novel view video, we report and compare the scores of ours and Generative Camera Dolly~\cite{van2024generative} on the VBench~\cite{huang2024vbench} benchmark. As we are primarily interested in video quality, we report the scores of 6 different metrics from the video quality assessment category. Specifically, we evaluate the scores of subject consistency, dynamic degree, motion smoothness, imaging quality, aesthetic quality, and background consistency. Further details of each metric is as follows:

\begin{itemize}
    \item \textbf{Subject Consistency:} The consistency of the subject in the video is calculated by DINO~\cite{caron2021emerging} feature similarity among frames. A higher score indicates higher consistency.
    \item \textbf{Dynamic Degree:} The dynamic degree calculates how much dynamics are included in the video, calculated by RAFT~\cite{teed2020raft}. A higher score indicates a more dynamic video.
    \item \textbf{Motion Smoothness:} Evaluates how smooth the motion in the video occurs, calculated by the motion priors in video frame interpolation models~\cite{li2023amt}. A higher score indicates smoother motions.
    \item \textbf{Imaging Quality:} Evaluates the quality of the individual frames that are synthesized. The scores are calculated by MUSIQ~\cite{ke2021musiq} image quality predictor. A higher score indicates higher quality with less distortion (\eg noise, blur).
    \item \textbf{Aesthetic Quality:} Evaluates the aesthetic quality of the individual frames that are synthesized. The aesthetic quality is aligned with human perception, calculated by the LAION~\cite{schuhmann2022laion} aesthetic predictor. A higher score indicates better richness and harmony of colors, more photo-realistic and natural.
    \item \textbf{Background Consistency:} The consistency of the background is calculated by the CLIP~\cite{radford2021learning} feature similarity across frames. A higher score indicates higher consistency.
\end{itemize}

\paragraph{Experimental setups for baselines.}
In Tab.~\ref{tab:main_quan}, we compare our method with baselines that render novel views of dynamic scenes without scene-specific optimization. For MonST3R~\cite{zhang2024monst3r}, our baseline, we evaluate two configurations. The first uses dynamic object masks obtained from global alignment process, incorporating pointmaps from all frames except dynamic regions, while the $t$-th frame includes both static and dynamic pointmaps. The second, MonST3R (Per-frame proj.), uses only pointmaps from the $t$-th frame. For novel view rendering, we input source and target frames to MonST3R to estimate relative poses and apply geometric warping to project pointmaps to the target viewpoint.

Next, we include Pseudo-DVS~\cite{zhao2024pseudo}, a reconstruction-based approach. As Pseudo-DVS predicts depth, optical flow, and camera poses as inputs during preprocessing, we ensure a fair comparison by normalizing and scaling the novel view camera poses to align with the depths and poses estimated by Pseudo-DVS from the multi-view video dataset. 

Finally, we evaluate against the generative approach Generative Camera Dolly (GCD)~\cite{van2024generative}. To evaluate the performance of GCD, we utilize the provided checkpoint of GCD trained on Kubric-4D~\cite{greff2022kubric} including movements up to 180 degrees, as this shows the best generalizability among the provided checkpoints. As the camera pose of Neu3D~\cite{li2022neural} and ST-NeRF~\cite{zhang2021editable} datasets have different scales with Kubric-4D, we have pre-processed the camera poses of Neu3D and ST-NeRF to Kubric-4D for fair comparison. We found that this pre-processing is essential for GCD as the scenes of Kubric-4D have a fixed configuration of the scene being in the middle of a sphere of $\text{radius} = 15$ and the cameras are located on the surface of the sphere. 

\section{Limitations}
\label{sec:appendix_limitations}
While our method effectively refines inaccurate predictions from geometry estimation models by leveraging the priors inherent in the generative model, it may exhibit limited performance in certain corner cases where the underlying geometry is severely mispredicted. Nevertheless, given that our approach is orthogonal to video geometry estimation, we believe our approach will continue to improve alongside the rapid advancements in geometry estimation models.

\section{Societal Impacts}
\label{sec:appendix_impact}
Our method, which enables the generation of videos from different angles based on the user's input video, has a potential risk of being misused for unintended purposes, such as the creation of inappropriate or unethical content. In such cases, as mentioned in \cite{hu2024animate}, methods for detecting AI-generated videos~\cite{wang2020deep,yu2020searching} could be employed.

\end{document}